\documentclass[preprint,11pt,3p,number]{elsarticle}

\usepackage{amsmath,amssymb}

\usepackage{algorithm}
\usepackage{algpseudocode}
\algrenewcommand\algorithmicrequire{\textbf{Input:}}
\algrenewcommand\algorithmicensure{\textbf{Output:}}

\usepackage{graphicx}
\usepackage{booktabs}
\usepackage{longtable}
\usepackage{tabularx}
\usepackage{subcaption}
\usepackage{array}

\usepackage{float}
\usepackage{placeins}
\usepackage{setspace}
\usepackage{chngcntr}

\usepackage{appendix}

\usepackage{xcolor}

\biboptions{sort&compress}

\usepackage[colorlinks=true,
            linkcolor=blue,
            citecolor=blue,
            urlcolor=blue]{hyperref}
\usepackage[nameinlink,capitalize]{cleveref}

\newcommand{\figref}[1]{Fig.~\ref{#1}}
\newcommand{\tabref}[1]{Table~\ref{#1}}

\newcommand{\eqnref}[1]{Eq.~\eqref{#1}}

\begin{document}
\doublespacing

\begin{frontmatter}

\title{Physics-Informed Discovery of Yield Functions in Plasticity via Convex Neural Representations}

\author[label1]{Hyeonbin Moon}
\author[label1]{Donghyuk Cho}
\author[label1]{Jecheon Yu}
\author[label1,label3]{Jeong Whan Yoon}
\author[label1,label2,label4]{Seunghwa Ryu\corref{cor1}}

\cortext[cor1]{Corresponding author}
\ead{ryush@kaist.ac.kr}

\affiliation[label1]{
organization={Department of Mechanical Engineering, Korea Advanced Institute of Science and Technology (KAIST)},
city={Daejeon},
postcode={34141},
country={Republic of Korea}
}

\affiliation[label2]{
organization={KAIST InnoCORE PRISM-AI Center, Korea Advanced Institute of Science and Technology (KAIST)},
city={Daejeon},
postcode={34141},
country={Republic of Korea}
}

\affiliation[label3]{
organization={School of Engineering, Deakin University},
addressline={75 Pigdons Road},
city={Waurn Ponds},
state={Victoria},
postcode={3216},
country={Australia}
}

\affiliation[label4]{
organization={Department of AX, College of AI, Korea Advanced Institute of Science and Technology (KAIST)},
city={Daejeon},
postcode={34141},
country={Republic of Korea}
}

\begin{abstract}
\begin{singlespace}
Identifying anisotropic yield functions remains challenging since yielding is not directly observed in full-field mechanical measurements, directional calibration can require many loading directions, and selecting an appropriate analytical form is nontrivial. This study proposes a physics-informed framework for discovering yield functions from full-field displacement data and reaction force data, without stress observations, plastic strain measurements, direct yield surface data, or a prescribed parametric yield function. The framework identifies the yield function as a mechanically constrained constitutive component inside elastoplastic stress integration, rather than through direct stress-space supervision. The yield function is represented by a convex neural network that enforces convexity and positive homogeneity of degree one while imposing the assumed tension-compression symmetry, and this neural yield function is trained with a differentiable stress update and a physics-informed force equilibrium loss across multiple loading cases. The proposed framework is validated using finite element (FE) benchmark studies with von Mises, Hill 1948, and Yld2000-2d yield functions, assessing yield contour agreement, displacement-noise sensitivity, identifiability through plastically active stress states, epistemic uncertainty, and polynomial-surrogate deployment. This study provides a mechanics-constrained pathway for discovering anisotropic yield functions from displacement and force data while keeping the identified component within the structure of elastoplastic stress integration.
\end{singlespace}
\end{abstract}


\end{frontmatter}

\section{Introduction}
\label{sec:introduction}

Constitutive modeling is fundamental to solid mechanics because it establishes material response relations for interpreting and predicting deformation and stress. In plasticity, this relation is governed by the yield function, hardening law, and flow rule. Among these components, the yield function is central to plasticity modeling because it specifies the elastic domain in stress space, determines the onset of plastic deformation, and, under associated flow, controls the direction of plastic straining. Classical yield functions such as von Mises and Tresca provide standard descriptions for isotropic yielding, while Hill-type and Barlat--Yoon-type yield functions extend this structure to anisotropic sheet metals and textured materials \cite{SimoHughes1998,Hill1948,Barlat2003Yld2000}. These analytical forms are interpretable and widely used, but their predictive accuracy depends on selecting a functional form that is sufficiently expressive for the material and stress states of interest.

Data-driven constitutive modeling has emerged as an alternative by replacing selected analytical constitutive assumptions with data-driven stress-response or material representations. Neural constitutive models can approximate nonlinear and history-dependent stress responses, and mechanics-informed variants embed thermodynamic restrictions, modular elastoplastic structure, or physics-informed training losses \cite{KirchdoerferOrtiz2016,Raissi2019PINN,Karniadakis2021PIML,Masi2021TANN,VlassisSun2021SobolevPlasticity,MeyerEkre2023Plasticity,Fuhg2023ModularElastoplasticity}. Related developments have introduced invariant, probabilistic, sparse, and physics-constrained material models for solid mechanics \cite{HeiderWangSun2020SO3,FuhgBouklas2022ProbabilisticConstitutive,LinkaKuhl2023CANN,Linden2023NeuralHyperelasticity,FuhgJonesBouklas2024Sparsification}. These approaches increase expressive capacity, especially when the constitutive response is too complex to be represented by a small set of prescribed parameters. However, expressive stress-response models do not necessarily reveal the constitutive structure that controls yielding and plastic deformation in conventional elastoplastic updates.

Constitutive discovery methods further improve interpretability by identifying constitutive structure rather than only stress histories. Symbolic regression can produce closed-form expressions from data \cite{SchmidtLipson2009,Brunton2016SINDy,Bomarito2021SymbolicPlasticity,Kabliman2021SymbolicPlasticity}, and sparse-regression frameworks such as EUCLID identify constitutive models from full-field displacement and reaction force data without direct stress measurements \cite{Flaschel2021HyperelasticDiscovery,Flaschel2022Plasticity,Joshi2022BayesianEUCLID,Flaschel2023EUCLID,Thakolkaran2022NNEUCLID}. Other machine-learning methods address yield surface prediction or stress integration in anisotropic plasticity using stress-level, crystal-plasticity, or stress-update data as supervision \cite{Nascimento2023YieldSurfaces,FazilyYoon2023StressIntegration,Fuhg2023EnhancingYield}. These approaches rely on different sources of supervision and different constitutive assumptions, ranging from displacement and force data to candidate libraries and stress-space supervision.

For plasticity, however, identifying the yield function itself from experimentally accessible displacement and force data remains challenging. Full-field displacement data are commonly obtained from digital image correlation (DIC) and provide spatially resolved deformation measurements, but they provide neither stress fields nor yield surface data directly \cite{Sutton2009DIC,Avril2008FullField}. Existing inverse-identification methods that use full-field displacement measurements can identify constitutive laws without direct stress measurements, but they often rely on prescribed candidate libraries when closed-form expressions are sought \cite{Flaschel2022Plasticity,Flaschel2023EUCLID}. Supervised neural approaches for yield surfaces or stress integration address related plasticity tasks, but they typically require stress-space data, crystal-plasticity data, or stress-update data \cite{Nascimento2023YieldSurfaces,FazilyYoon2023StressIntegration,Fuhg2023EnhancingYield}. This difficulty becomes more pronounced for anisotropic plasticity, where yielding can have strong directional dependence and where selecting a compact closed-form expression that remains expressive, convex, and compatible with stress integration is difficult \cite{Hill1948,Barlat2003Yld2000,Soare2008PolynomialYield,SoareBarlat2010ConvexPolynomial}. At the same time, a yield function used in elastoplastic stress integration needs to preserve the mechanical structure of the elastic domain and hardening representation. These limitations motivate a physics-informed framework that can infer a yield function from measured displacement and reaction force data while preserving the structure required for elastoplastic stress updates.

\begin{figure}[!htbp]
    \centering
    \includegraphics[width=\textwidth]{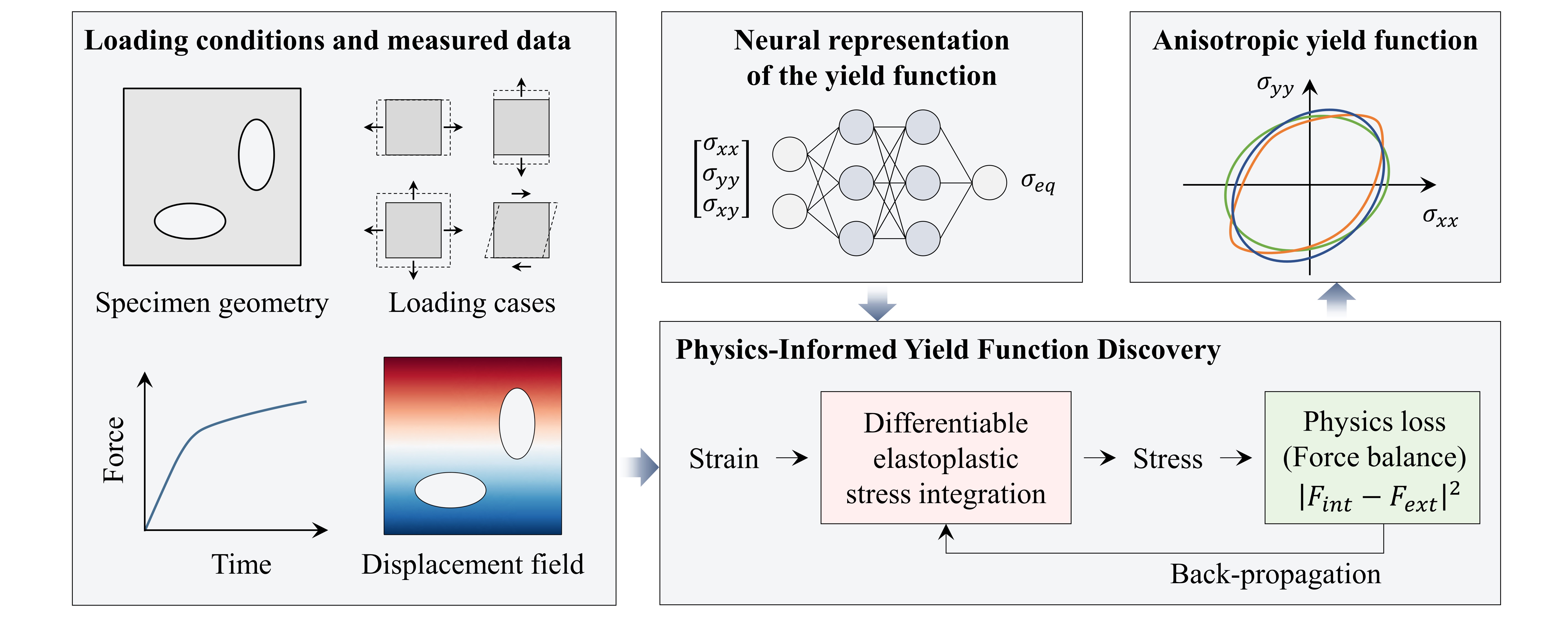}
    \caption{
    Schematic overview of the proposed physics-informed yield function discovery framework.
    Full-field displacement data and reaction force data from multiple loading cases are used to train a convex neural network representation of the yield function.
    Differentiable elastoplastic stress integration and force equilibrium residual minimization enable inference of the yield function from measured displacement and reaction force data in the prescribed benchmark setting.
    }
    \label{fig:framework_overview}
\end{figure}

This study proposes a physics-informed framework for discovering yield functions from full-field displacement data and reaction force data obtained under four loading cases. \figref{fig:framework_overview} illustrates the framework from displacement and force data to the discovered yield function. The unknown yield function is first represented by a convex neural network, with convexity and positive homogeneity of degree one enforced by construction and tension-compression symmetry imposed as the material assumption adopted in this study. Strain fields obtained from the displacement data are then used in an elastoplastic stress update, and the resulting stress fields are assembled into force predictions. The neural yield function is trained by minimizing force equilibrium residuals against the measured reaction force data, rather than by using direct supervision in stress space. A differentiable elastoplastic training algorithm embeds the neural yield function in the stress update so that the neural yield function is optimized using the force equilibrium residual. The framework is validated using data generated by finite element (FE) simulations from three benchmark yield functions, namely von Mises, Hill 1948, and Yld2000-2d. The validation considers the inferred yield contours, directional anisotropy, noise sensitivity, epistemic uncertainty, and finite element deployment through a smooth polynomial-type surrogate. The main contributions of this work are as follows.
\begin{itemize}
\item The proposed framework identifies yield functions from full-field displacement and reaction force data through force equilibrium, without stress observations, plastic strain measurements, direct yield surface data, or a prescribed analytical yield function.
\item A convex neural yield function representation and differentiable elastoplastic training algorithm are introduced to enforce key mechanical constraints (convexity of the yield function and positive homogeneity of degree one) while enabling optimization based on force equilibrium.
\item The validation study evaluates noise-free identification, displacement-noise sensitivity, epistemic uncertainty, and finite element deployment through a smooth polynomial-type surrogate.
\end{itemize}

The remainder of this paper is organized as follows. Section~\ref{sec:methodology} describes the problem setting and methodology, Section~\ref{sec:results} presents the numerical validation results, Section~\ref{sec:discussion} discusses implications and limitations, and Section~\ref{sec:conclusions} concludes the paper.

\FloatBarrier

\section{Methodology}
\label{sec:methodology}

\subsection{Problem Setting and Available Data}
\label{subsec:problem_setting}

\begin{figure}[!htbp]
    \centering
    \includegraphics[width=\textwidth]{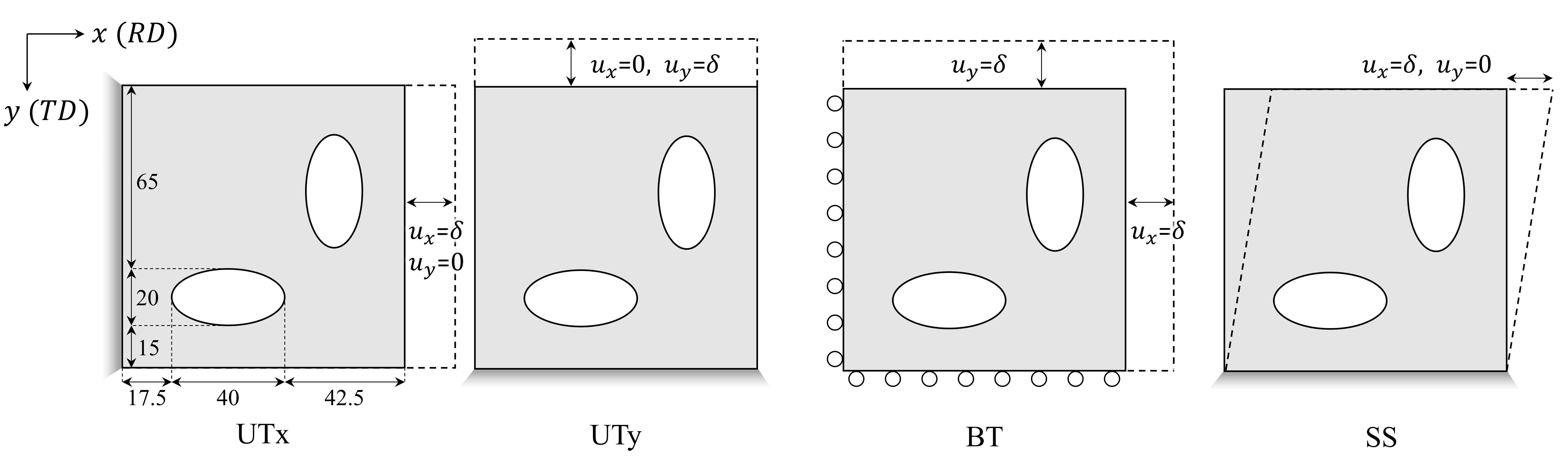}
    \caption{
    Specimen geometry and loading cases used for yield function discovery.
    A rectangular specimen with two elliptical voids is considered to induce heterogeneous stress states.
    Four loading cases are applied to the same geometry: uniaxial tension in the $x$-direction (UTx), uniaxial tension in the $y$-direction (UTy), biaxial tension (BT), and simple shear (SS).
    The $x$- and $y$-directions correspond to the rolling direction (RD) and transverse direction (TD), respectively.
    }
    \label{fig:problem_setting}
\end{figure}

Consider a rectangular specimen with two elliptical voids under a two-dimensional plane-stress condition, subjected to four loading cases on the same geometry, as shown in \figref{fig:problem_setting}: uniaxial tension in the $x$-direction (UTx), uniaxial tension in the $y$-direction (UTy), biaxial tension (BT), and simple shear (SS) \cite{Flaschel2022Plasticity,Xu2025EUCLIDPressureSensitive}. The loading cases are defined through prescribed boundary displacements, while the remaining degrees of freedom are left unconstrained unless otherwise specified in \figref{fig:problem_setting}.
The $x$- and $y$-directions correspond to the rolling direction (RD) and transverse direction (TD), respectively. For each loading case, the available data are time-resolved full-field displacement data, as typically obtained from DIC, and reaction force data on the loaded boundaries. The aim is to identify a yield function shared across the available loading cases from these measured data.

The two-void geometry and the four loading cases are used to generate heterogeneous deformation fields and mixed stress states, so that the inverse problem is not reduced to fitting a yield function from homogeneous stress--strain responses. A spatially uniform stress state would constrain only a limited region of the yield surface, whereas the voids and multiple loading cases induce diverse stress states in the measured response. During discovery, stress fields, plastic strain fields, and direct yield surface data are not used as training data. Instead, displacement data are used to compute strain increments, while reaction force data provide global equilibrium constraints on the loaded boundaries. Here, the reaction force denotes the resultant force measured on the displacement-controlled boundary, corresponding to the global load response of the specimen. Using heterogeneous displacement fields is consistent with inverse-identification approaches in which nonuniform deformation fields constrain constitutive behavior \cite{Avril2008FullField,Lourenco2024IndirectVFM,WuZhangMao2025PDEConstrained,TungLi2024AntiDogbone}.

In this study, the material response is described within small-deformation, rate-independent, pressure-insensitive plasticity. The material is assumed to have a known linear elastic stiffness, a known isotropic hardening law, a prescribed associated flow rule (AFR), and an assumed tension-compression symmetry of the yield function. The hardening response is treated as a prescribed input; in an experimental application, it could be determined from an independent calibration and expressed in the form
\begin{equation}
    R(\bar{\varepsilon}^{p})
    =
    \mathcal{R}(\bar{\varepsilon}^{p};\boldsymbol{p}_{R}),
    \label{eq:hardening_law}
\end{equation}
where $\mathcal{R}$ denotes the prescribed hardening law, $\bar{\varepsilon}^{p}$ denotes the equivalent plastic strain, and $\boldsymbol{p}_{R}$ denotes the hardening parameters.

\FloatBarrier

\subsection{Neural Representation of the Yield Function}
\label{subsec:nn_yield_function}

The unknown yield function is represented by a neural network instead of being selected from a prescribed parametric yield function or candidate library. Before specifying the neural network architecture, the mechanical constraints on the yield function are stated.

Let $\boldsymbol{\sigma}=[\sigma_{xx},\sigma_{yy},\sigma_{xy}]^{T}$ denote the in-plane stress vector. Under the rate-independent, pressure-insensitive plasticity setting considered here, yielding is defined by separating the yield function $\bar{\sigma}(\boldsymbol{\sigma})$ from the hardening response as
\begin{equation}
    f(\boldsymbol{\sigma},\bar{\varepsilon}^{p})
    =
    \bar{\sigma}(\boldsymbol{\sigma})
    -
    R(\bar{\varepsilon}^{p})
    \leq 0 .
    \label{eq:yield_condition}
\end{equation}
In a plastic potential formulation, stability analyses for metal plasticity place separate positivity constraints on the stress projections of the derivatives of the plastic potential and the yield function,
\begin{equation}
    \left[
    \partial_{\boldsymbol{\sigma}}\bar{\psi}(\boldsymbol{\sigma})
    \right]^{T}
    \boldsymbol{\sigma}>0,
    \qquad
    \left[
    \partial_{\boldsymbol{\sigma}}\bar{\sigma}(\boldsymbol{\sigma})
    \right]^{T}
    \boldsymbol{\sigma}>0,
    \label{eq:stoughton_yoon_constraints}
\end{equation}
where $\bar{\psi}$ denotes the plastic potential. The first inequality ensures positive plastic work, whereas the second is imposed on the yield function to avoid instabilities associated with yield point elongation \cite{StoughtonYoon2006Drucker}. For the AFR adopted here, $\bar{\psi}=\bar{\sigma}$ and the plastic strain increment is
\begin{equation}
    \Delta\boldsymbol{\varepsilon}^{p}
    =
    \Delta\bar{\varepsilon}^{p}
    \partial_{\boldsymbol{\sigma}}\bar{\sigma}(\boldsymbol{\sigma}) ,
    \label{eq:associated_flow}
\end{equation}
where $\Delta\bar{\varepsilon}^{p}$ is the equivalent plastic strain increment and $\Delta\boldsymbol{\varepsilon}^{p}$ is the plastic strain increment. The plastic work increment under the AFR is
\begin{equation}
    \Delta W^{p}
    =
    \boldsymbol{\sigma}^{T}\Delta\boldsymbol{\varepsilon}^{p}
    =
    \Delta\bar{\varepsilon}^{p}
    \left[
    \partial_{\boldsymbol{\sigma}}\bar{\sigma}(\boldsymbol{\sigma})
    \right]^{T}
    \boldsymbol{\sigma}.
    \label{eq:plastic_work_increment}
\end{equation}
Since $\Delta\bar{\varepsilon}^{p}$ is positive during plastic deformation, positivity of plastic work is governed by the sign of the stress projection of the derivative of the yield function with respect to stress in \eqnref{eq:plastic_work_increment}. The AFR also makes the plastic strain increment normal to the yield surface. For rate-independent associated plasticity, a convex elastic domain is consistent with the standard stability structure associated with normality, the maximum-plastic-work interpretation, and Drucker's stability postulate \cite{Hill1950Plasticity,Hill1958Stability,Franchi1990QuasiConvexity,StoughtonYoon2006Drucker,SimoHughes1998}. The yield function $\bar{\sigma}$ is therefore constrained to be convex. Under the AFR, the two constraints in \eqnref{eq:stoughton_yoon_constraints} reduce to the same condition, $\left[\partial_{\boldsymbol{\sigma}}\bar{\sigma}(\boldsymbol{\sigma})\right]^{T}\boldsymbol{\sigma}>0$, which is the positivity condition associated with \eqnref{eq:plastic_work_increment}. Choosing $\bar{\psi}$ and $\bar{\sigma}$ to be positively homogeneous of degree one provides a sufficient way to satisfy these constraints. For functions with this homogeneity, the stress projection of the derivative equals the function value, so the two terms in \eqnref{eq:stoughton_yoon_constraints} reduce to $\bar{\psi}(\boldsymbol{\sigma})$ and $\bar{\sigma}(\boldsymbol{\sigma})$, respectively. The constraints themselves are more general and can also be satisfied by certain non-homogeneous functions. Positive homogeneity of degree one is therefore adopted here as a sufficient restriction for the pressure-insensitive metal-plasticity setting considered in this study, not as a general necessity of plasticity. Because the AFR uses $\partial_{\boldsymbol{\sigma}}\bar{\sigma}$, smoothness is also relevant, although nonsmooth convex yield surfaces are established in plasticity \cite{BigoniPiccolroaz2004YieldCriteria,PiccolroazBigoni2009Corners,Lubliner1990Plasticity,SimoHughes1998}. The convexity and homogeneity conditions imposed on $\bar{\sigma}$ are therefore expressed as
\begin{align}
    \bar{\sigma}
    \left(\lambda\boldsymbol{\sigma}_{1}+(1-\lambda)\boldsymbol{\sigma}_{2}\right)
    &\leq
    \lambda\bar{\sigma}(\boldsymbol{\sigma}_{1})
    +(1-\lambda)\bar{\sigma}(\boldsymbol{\sigma}_{2}),
    \qquad \lambda\in[0,1], \label{eq:convexity_requirement}\\
    \bar{\sigma}(\rho\boldsymbol{\sigma})
    &=
    \rho\bar{\sigma}(\boldsymbol{\sigma}),
    \qquad \rho>0 . \label{eq:homogeneity_requirement}
\end{align}

\begin{figure}[!htbp]
    \centering
    \includegraphics[width=\textwidth]{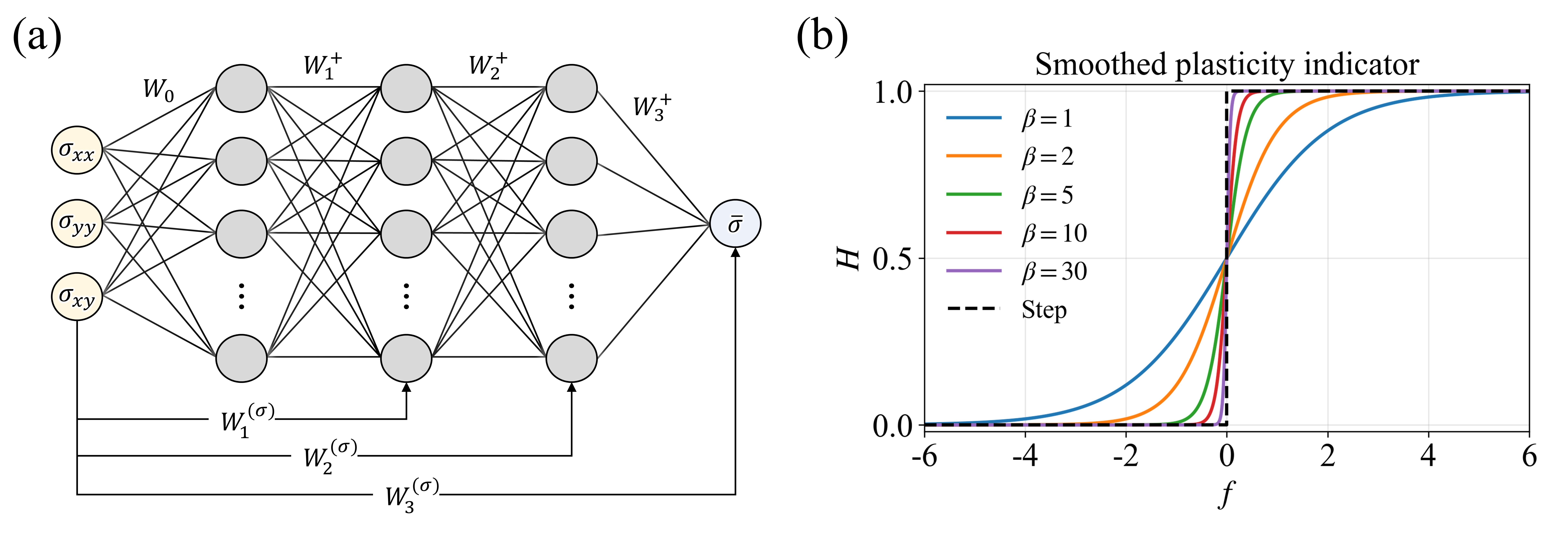}
    \caption{
    Convex neural network architecture and smoothed plasticity indicator.
    (a) Input convex neural network structure used for the neural yield function.
    (b) Sigmoid approximation of the plasticity indicator with different sharpness parameters $\beta$, introduced in Section~\ref{subsec:physics_informed_training} and used to enable differentiable stress integration during training.
    }
    \label{fig:cnn_and_smooth_indicator}
\end{figure}

The neural yield function is then constructed to enforce \eqnref{eq:convexity_requirement} and \eqnref{eq:homogeneity_requirement} while allowing evaluation of the derivative of the yield function with respect to stress required by the AFR. Global $C^{1}$ smoothness would make this derivative continuous, but to the best of the authors' knowledge, no neural network architecture simultaneously enforces convexity, positive homogeneity of degree one, and global $C^{1}$ smoothness. The neural network architecture adopted in this study therefore enforces convexity and positive homogeneity by construction while relaxing global $C^{1}$ smoothness to a $C^{0}$, piecewise differentiable representation. Convexity is enforced using an input convex neural network (hereafter referred to as a convex neural network) \cite{Amos2017ICNN}, whose architecture is illustrated in \figref{fig:cnn_and_smooth_indicator}(a). The convex neural network uses the in-plane stress vector $\boldsymbol{\sigma}$ as input and returns the scalar neural yield function $\bar{\sigma}_{\boldsymbol{\theta}}(\boldsymbol{\sigma})$, where $\boldsymbol{\theta}$ denotes the trainable parameters. The layer update is given by
\begin{align}
    \mathbf{z}_{1}
    &=
    \mathrm{ReLU}\left(\mathbf{W}_{x}^{(1)}\boldsymbol{\sigma}\right), \\
    \mathbf{z}_{\ell}
    &=
    \mathrm{ReLU}\left(
    \mathbf{W}_{x}^{(\ell)}\boldsymbol{\sigma}
    +
    g\left(\mathbf{W}_{z}^{(\ell)}\right)\mathbf{z}_{\ell-1}
    \right),
    \qquad \ell=2,\ldots,L, \label{eq:icnn_forward}
\end{align}
where the elementwise map
\begin{equation}
    g(\omega)=\sqrt{\omega^{2}+\gamma^{2}}-\gamma,
    \qquad \gamma=0.1,
    \label{eq:nonnegative_reparameterization}
\end{equation}
keeps the weights multiplying the previous-layer activation nonnegative in a differentiable manner. In \eqnref{eq:icnn_forward}, $\mathbf{z}_{\ell}$ denotes the hidden activation at layer $\ell$, $L$ is the number of hidden layers, $\mathbf{W}_{x}^{(\ell)}$ maps the stress input to layer $\ell$, and $\mathbf{W}_{z}^{(\ell)}$ maps the previous hidden activation to the current layer. These architectural choices are tied to the constraints on the yield function. ReLU activations preserve positive homogeneity of degree one, removing bias terms avoids additive offsets, and the nonnegative weights on the previous-layer activations preserve input convexity through the network composition. The smooth reparameterization in \eqnref{eq:nonnegative_reparameterization} enforces nonnegativity without a projection step during gradient-based optimization.

The assumed tension-compression symmetry is incorporated by symmetrizing the convex neural network output over the sign transformations used in \eqnref{eq:symmetric_yield}, namely simultaneous reversal of the normal stress components and independent reversal of the shear stress component.
\begin{equation}
    \bar{\sigma}_{\boldsymbol{\theta}}(\boldsymbol{\sigma})
    =
    \frac{1}{4}
    \sum_{\eta_{1},\eta_{2}\in\{-1,1\}}
    \mathrm{NN}_{\boldsymbol{\theta}}
    \left(
    \eta_{1}\sigma_{xx},
    \eta_{1}\sigma_{yy},
    \eta_{2}\sigma_{xy}
    \right),
    \label{eq:symmetric_yield}
\end{equation}
where $\mathrm{NN}_{\boldsymbol{\theta}}$ denotes the raw convex neural network output. The neural yield function $\bar{\sigma}_{\boldsymbol{\theta}}$ is constrained by convexity, the adopted positive homogeneity of degree one, and the assumed symmetry rather than by a selected analytical equation.

\FloatBarrier

\subsection{Physics-Informed Training Strategy}
\label{subsec:physics_informed_training}

The training procedure combines displacement-derived strain increments, elastoplastic stress integration, and force equilibrium residuals. For each loading case, the displacement field is interpolated on the finite element mesh, and nodal displacement increments are converted into strain increments through the standard FE strain-displacement relation
\begin{equation}
    \Delta \boldsymbol{\varepsilon}_{e}^{n}
    =
    \mathbf{B}_{e}\Delta \mathbf{u}_{e}^{n},
    \label{eq:strain_displacement}
\end{equation}
where $\Delta \mathbf{u}_{e}^{n}$ is the nodal displacement increment of element $e$ at time increment $n$, $\mathbf{B}_{e}$ is the element strain-displacement matrix, and $\Delta \boldsymbol{\varepsilon}_{e}^{n}=[\Delta\varepsilon_{xx},\Delta\varepsilon_{yy},\Delta\gamma_{xy}]^{T}$ under plane stress. Throughout the manuscript, in-plane strain vectors use the engineering shear strain convention, $\boldsymbol{\varepsilon}=[\varepsilon_{xx},\varepsilon_{yy},\gamma_{xy}]^{T}$.

The strain increments obtained from the displacement data are then used in an explicit forward Euler stress integration scheme. Given a strain increment $\Delta\boldsymbol{\varepsilon}^{n}$, the forward Euler update is expressed as
\begin{align}
    \Delta\boldsymbol{\sigma}^{n}
    &=
    \mathbf{C}_{e}
    \left(
    \Delta\boldsymbol{\varepsilon}^{n}
    -
    \Delta\bar{\varepsilon}^{p,n}
    \mathbf{n}^{n-1}
    \right),
    \label{eq:forward_euler_stress_increment}\\
    \Delta\bar{\varepsilon}^{p,n}
    &=
    \left\langle
    \frac{
    \left(\mathbf{n}^{n-1}\right)^{T}
    \mathbf{C}_{e}
    \Delta\boldsymbol{\varepsilon}^{n}
    }{
    \partial_{\bar{\varepsilon}^{p}}R
    +
    \left(\mathbf{n}^{n-1}\right)^{T}
    \mathbf{C}_{e}
    \mathbf{n}^{n-1}
    }
    H^{n}
    \right\rangle_{+},
    \label{eq:forward_euler_plastic_increment}\\
    \mathbf{n}^{n-1}
    &=
    \partial_{\boldsymbol{\sigma}}
    \bar{\sigma}
    \left(\boldsymbol{\sigma}^{n-1}\right)
    \label{eq:flow_direction}
\end{align}
where $\mathbf{n}^{n-1}$ is the derivative of the yield function with respect to stress used in the explicit plastic correction, $\mathbf{C}_{e}$ is the plane-stress elastic stiffness matrix, $\partial_{\bar{\varepsilon}^{p}}R$ is evaluated at $\bar{\varepsilon}^{p,n-1}$, and the nonnegative projection is defined by $\langle x\rangle_{+}=\max(x,0)$. The plasticity indicator $H^{n}$ is determined by the sign of the trial yield residual $f_{\mathrm{trial}}^{n}$ evaluated at the trial stress $\boldsymbol{\sigma}_{\mathrm{trial}}^{n}$,
\begin{align}
    H^{n}
    &=
    \begin{cases}
    0, & \text{if } f_{\mathrm{trial}}^{n}<0,\\
    1, & \text{if } f_{\mathrm{trial}}^{n}\geq 0,
    \end{cases}
    \label{eq:plasticity_indicator}\\
    f_{\mathrm{trial}}^{n}
    &=
    f\!\left(
    \boldsymbol{\sigma}_{\mathrm{trial}}^{n},
    \bar{\varepsilon}^{p,n-1}
    \right)
    =
    \bar{\sigma}\!\left(\boldsymbol{\sigma}_{\mathrm{trial}}^{n}\right)
    -
    R(\bar{\varepsilon}^{p,n-1}).
    \label{eq:trial_yield_residual}\\
    \boldsymbol{\sigma}_{\mathrm{trial}}^{n}
    &=
    \boldsymbol{\sigma}^{n-1}
    +
    \mathbf{C}_{e}\Delta\boldsymbol{\varepsilon}^{n}.
    \label{eq:trial_stress}
\end{align}
Thus, $H^{n}=0$ corresponds to an elastic increment and $H^{n}=1$ activates the plastic correction.

The forward Euler update above provides the stress increment employed in the training procedure. In this study, $\bar{\sigma}$ is represented by $\bar{\sigma}_{\boldsymbol{\theta}}$, so the update is required to remain differentiable with respect to the trainable yield function parameters. The indicator $H^{n}$ changes abruptly from 0 to 1 at the yield boundary, and this step change is not differentiable. For training, the trial yield residual is evaluated with the neural yield function,
\begin{equation}
    f_{\boldsymbol{\theta},\mathrm{trial}}^{n}
    =
    \bar{\sigma}_{\boldsymbol{\theta}}
    \left(\boldsymbol{\sigma}_{\mathrm{trial}}^{n}\right)
    -
    R(\bar{\varepsilon}^{p,n-1}),
    \label{eq:neural_trial_yield_residual}
\end{equation}
and $H^{n}$ is replaced by the smooth plasticity indicator
\begin{equation}
    H_{\beta}^{n}
    =
    \left[
    1+
    \exp\left\{
    -\beta
    f_{\boldsymbol{\theta},\mathrm{trial}}^{n}
    \right\}
    \right]^{-1},
    \label{eq:smooth_indicator}
\end{equation}
where $\beta$ is a sharpness parameter for the sigmoid approximation. \figref{fig:cnn_and_smooth_indicator}(b) illustrates that larger $\beta$ values sharpen the transition. The derivative of the neural yield function with respect to stress used in the differentiable update is $\mathbf{n}_{\boldsymbol{\theta}}^{n-1}=\partial_{\boldsymbol{\sigma}}\bar{\sigma}_{\boldsymbol{\theta}}(\boldsymbol{\sigma}^{n-1})$. With this replacement, the differentiable plastic increment is computed as
\begin{equation}
    \Delta\bar{\varepsilon}^{p,n}
    =
    \left\langle
    \frac{
    \left(\mathbf{n}_{\boldsymbol{\theta}}^{n-1}\right)^{T}
    \mathbf{C}_{e}
    \Delta\boldsymbol{\varepsilon}^{n}
    }{
    \partial_{\bar{\varepsilon}^{p}}R
    +
    \left(\mathbf{n}_{\boldsymbol{\theta}}^{n-1}\right)^{T}
    \mathbf{C}_{e}
    \mathbf{n}_{\boldsymbol{\theta}}^{n-1}
    }
    H_{\beta}^{n}
    \right\rangle_{+},
    \label{eq:plastic_increment}
\end{equation}
with the same definitions of $\partial_{\bar{\varepsilon}^{p}}R$ and $\langle\cdot\rangle_{+}$ as in \eqnref{eq:forward_euler_plastic_increment}.
The corresponding stress increment is
\begin{equation}
    \Delta\boldsymbol{\sigma}^{n}
    =
    \mathbf{C}_{e}
    \left(
    \Delta\boldsymbol{\varepsilon}^{n}
    -
    \Delta\bar{\varepsilon}^{p,n}
    \mathbf{n}_{\boldsymbol{\theta}}^{n-1}
    \right).
    \label{eq:stress_increment}
\end{equation}
This update preserves differentiability of the stress response with respect to the neural yield function while avoiding an implicit return-mapping solve during training. The explicit forward Euler scheme is used because an implicit backward Euler return-mapping update would require Newton iterations and the associated consistent linearization. For a neural yield function, that linearization involves second derivatives of the yield function with respect to stress, which are not well defined globally for the $C^{0}$, piecewise differentiable representation used here.

For each loading case $c$ and time increment $n=1,\ldots,N_{t}$, the force equilibrium loss is evaluated from the difference between an internal force-increment vector predicted from the stress update and an external force-increment vector constructed from the measured reactions and zero-force equilibrium conditions. At the element level, the internal force increment is
\begin{equation}
    \Delta\mathbf{F}_{\mathrm{int},e}^{n}
    =
    \int_{\Omega_{e}}
    \mathbf{B}_{e}^{T}
    \Delta\boldsymbol{\sigma}_{e}^{n}
    \,d\Omega ,
    \label{eq:internal_force_increment}
\end{equation}
where $\Omega_{e}$ is the domain of element $e$. The integral is evaluated numerically over the integration points. The element-level force increments are assembled into the global nodal force vector for loading case $c$,
\begin{equation}
    \Delta\mathbf{F}_{\mathrm{int,glob},c}^{n}
    =
    \mathcal{A}_{e}
    \left[
    \Delta\mathbf{F}_{\mathrm{int},e}^{n}
    \right],
    \label{eq:global_internal_force_increment}
\end{equation}
where the bracketed term denotes the collection of element force increments over the mesh, and $\mathcal{A}_{e}$ is the finite element assembly operator that scatters the element contributions into the global nodal force vector. The loading-case vector used in the residual is then
\begin{equation}
    \Delta\mathbf{F}_{\mathrm{int},c}^{n}
    =
    \mathbf{L}_{c}
    \Delta\mathbf{F}_{\mathrm{int,glob},c}^{n},
    \label{eq:boundary_force_aggregation}
\end{equation}
where $\mathbf{L}_{c}$ maps the global vector to the degrees of freedom used for loading case $c$. These entries consist of reaction-force components on prescribed-displacement boundaries and selected unconstrained degrees of freedom where the external-force increment is zero. The corresponding external vector $\Delta\mathbf{F}_{\mathrm{ext},c}^{n}$ is constructed from the measured reaction force increments for the reaction-force entries and zero entries for the unconstrained equilibrium entries. The case-wise force equilibrium loss is then
\begin{equation}
    \mathcal{L}_{c}(\boldsymbol{\theta})
    =
    \sum_{n=1}^{N_{t}}
    \left\|
    \Delta\mathbf{F}_{\mathrm{int},c}^{n}(\boldsymbol{\theta})
    -
    \Delta\mathbf{F}_{\mathrm{ext},c}^{n}
    \right\|_{2}^{2},
    \label{eq:case_loss}
\end{equation}
The loss in \eqnref{eq:case_loss} therefore penalizes the mismatch between the internal force increments predicted from the neural yield function stress update and the external force increments obtained from the reaction force data and zero-force equilibrium conditions. The total training loss combines the loading cases as
\begin{equation}
    \mathcal{L}(\boldsymbol{\theta})
    =
    \sum_{c\in\{\mathrm{UTx},\mathrm{UTy},\mathrm{BT},\mathrm{SS}\}}
    w_{c}\mathcal{L}_{c}(\boldsymbol{\theta}),
    \label{eq:total_loss}
\end{equation}
where $w_{c}$ denotes the weight assigned to loading case $c$. The same parameters $\boldsymbol{\theta}$ are shared across all loading cases, while the displacement histories, element matrices, elastic stiffness, and hardening law remain fixed. The parameters are optimized with a gradient-based optimizer, and the force equilibrium residual provides the physics-informed training signal without stress field measurements, plastic strain measurements, or direct yield surface data.

The differentiable stress update introduces a memory constraint in the implementation. After the initial elastic starting step, the smooth indicator $H_{\beta}$ depends on the neural yield function evaluation at each plastic-correction increment and remains differentiable with respect to the trainable parameters. The explicit plastic correction also uses the flow direction $\mathbf{n}_{\boldsymbol{\theta}}=\partial_{\boldsymbol{\sigma}}\bar{\sigma}_{\boldsymbol{\theta}}$. Differentiating this direction with respect to the neural network parameters introduces higher-order derivatives, and propagating these terms through every element, integration point, loading case, and time increment would require substantial memory. In the current implementation, every sampled increment advances the stress and internal-variable histories and contributes to the force equilibrium residual, while differentiation through $\mathbf{n}_{\boldsymbol{\theta}}$ with respect to $\boldsymbol{\theta}$ is retained only at selected increments.

Let $\mathcal{G}_{c}\subset\{1,\ldots,N_{t}\}$ denote the increments at which differentiation of $\mathbf{n}_{\boldsymbol{\theta}}$ with respect to $\boldsymbol{\theta}$ is retained for loading case $c$. The set $\mathcal{G}_{c}$ does not change the force equilibrium loss in \eqnref{eq:case_loss}: the residual is still accumulated over all sampled increments. At increments outside $\mathcal{G}_{c}$, $\mathbf{n}_{\boldsymbol{\theta}}$ is evaluated for the forward stress update but treated as fixed with respect to $\boldsymbol{\theta}$ in the optimization gradient. These increments still contribute to training through the differentiable indicator $H_{\beta}$ and the force equilibrium residual. If sufficient memory were available, $\mathcal{G}_{c}$ could include all sampled increments.

Algorithm~\ref{alg:training} summarizes the training procedure, including conversion of displacement increments to strain increments, differentiable forward Euler stress update, force equilibrium residual evaluation, and the selected increments at which differentiation of the derivative of the neural yield function with respect to stress is retained. In the algorithm, $k$ denotes the training epoch, $c$ denotes the loading case, and $n$ denotes the time increment. The set $\mathcal{C}$ contains the loading cases included in the inverse problem, and $\mathcal{D}_{c}$ denotes the displacement and reaction force data for case $c$. The first sampled loading increment is initialized elastically as a numerical starting step, and the smoothed plastic-correction steps are then applied to the subsequent increments. The sharpness parameter $\beta_{k}$ used in Algorithm~\ref{alg:training} is updated by Algorithm~\ref{alg:sharpness_schedule}. The schedule uses $\zeta=1/\beta$ as the transition width of the sigmoid indicator. Starting from $\zeta_{0}=1/\beta_{0}$, the transition width is multiplied by the factor $\eta_{\zeta}$ every $N_{\zeta}$ epochs.

\begin{algorithm}[!htbp]
\caption{Physics-informed training of the neural yield function}
\label{alg:training}
\begin{algorithmic}[1]
\Statex \textbf{Input:} Data $\{\mathcal{D}_{c}\}_{c\in\mathcal{C}}$, selected increments $\{\mathcal{G}_{c}\}_{c\in\mathcal{C}}$, and initial $\boldsymbol{\theta}$
\Statex \hspace{\algorithmicindent} Fixed quantities $\mathbf{C}_{e}$, $R(\bar{\varepsilon}^{p})$, $\mathbf{B}_{e}$, and $\mathbf{L}_{c}$
\Statex \textbf{Output:} Identified neural yield function $\bar{\sigma}_{\boldsymbol{\theta}}$
\State Initialize $\bar{\sigma}_{\boldsymbol{\theta}}$ and the optimization state
\For{epoch $k=1,\ldots,N_{\mathrm{epoch}}$}
    \State Evaluate $\beta_{k}$ using Algorithm~\ref{alg:sharpness_schedule}
    \State Initialize $\mathcal{L}\gets0$
    \For{each loading case $c\in\mathcal{C}$}
        \State Compute strain increments $\Delta\boldsymbol{\varepsilon}_{e}^{n}=\mathbf{B}_{e}\Delta\mathbf{u}_{e}^{n}$ from $\mathcal{D}_{c}$
        \State Initialize $\boldsymbol{\sigma}_{e}^{0}\gets\mathbf{0}$ and $\bar{\varepsilon}_{e}^{p,0}\gets0$ at all integration points
        \For{each time increment $n$}
            \State Compute $\boldsymbol{\sigma}_{\mathrm{trial}}^{n}=\boldsymbol{\sigma}^{n-1}+\mathbf{C}_{e}\Delta\boldsymbol{\varepsilon}^{n}$
            \If{$n$ is the first sampled increment}
                \State Set $\Delta\bar{\varepsilon}^{p,n}\gets0$ and $\Delta\boldsymbol{\sigma}^{n}\gets\mathbf{C}_{e}\Delta\boldsymbol{\varepsilon}^{n}$
            \Else
                \State Evaluate $H_{\beta_{k}}^{n}$ from \eqnref{eq:smooth_indicator}
                \State Compute the local derivatives $\mathbf{n}_{\boldsymbol{\theta}}^{n-1}$ and $\partial_{\bar{\varepsilon}^{p}}R$
                \If{$n\notin\mathcal{G}_{c}$}
                    \State Detach $\mathbf{n}_{\boldsymbol{\theta}}^{n-1}$ from the differentiation graph
                \EndIf
                \State Compute $\Delta\bar{\varepsilon}^{p,n}$ and $\Delta\boldsymbol{\sigma}^{n}$ from \eqnref{eq:plastic_increment} and \eqnref{eq:stress_increment}
            \EndIf
            \State Update the stress and internal-variable histories
            \State $\boldsymbol{\sigma}^{n}\gets\boldsymbol{\sigma}^{n-1}+\Delta\boldsymbol{\sigma}^{n}$
            \State $\bar{\varepsilon}^{p,n}\gets\bar{\varepsilon}^{p,n-1}+\Delta\bar{\varepsilon}^{p,n}$
            \State Assemble $\Delta\mathbf{F}_{\mathrm{int},c}^{n}$ from \eqnref{eq:boundary_force_aggregation}
            \State Accumulate $\mathcal{L}\gets\mathcal{L}+w_{c}\left\|\Delta\mathbf{F}_{\mathrm{int},c}^{n}-\Delta\mathbf{F}_{\mathrm{ext},c}^{n}\right\|_{2}^{2}$
        \EndFor
    \EndFor
    \State Evaluate the gradient of $\mathcal{L}$ and update $\boldsymbol{\theta}$
\EndFor
\end{algorithmic}
\end{algorithm}

\begin{algorithm}[!htbp]
\caption{Sharpness schedule for the smooth plasticity indicator}
\label{alg:sharpness_schedule}
\begin{algorithmic}[1]
\Statex \textbf{Input:} Epoch $k$, initial sharpness $\beta_{0}$, update factor $\eta_{\zeta}$, and update period $N_{\zeta}$
\Statex \textbf{Output:} Sharpness parameter $\beta_{k}$
\State Initialize the transition width as $\zeta_{0}=1/\beta_{0}$
\State Compute the schedule index $m_{k}=\left\lfloor(k-1)/N_{\zeta}\right\rfloor$
\State Update the transition width as $\zeta_{k}=\eta_{\zeta}^{m_{k}}\zeta_{0}$
\State Return $\beta_{k}=1/\zeta_{k}$
\end{algorithmic}
\end{algorithm}

\FloatBarrier

\subsection{Polynomial Surrogate for Finite Element Deployment}
\label{subsec:polynomial_surrogate}

The convex neural network representation is designed for yield function discovery, whereas conventional implicit finite element deployment requires additional properties for the constitutive function used in return mapping. In particular, the neural yield function is treated as $C^{0}$ and piecewise differentiable, while an implicit return-mapping implementation requires a smooth analytical representation for stable implementation, with derivatives that can be evaluated consistently. The discovered neural yield function is therefore post-processed into a smooth polynomial-type surrogate for deployment, after the inverse identification step has been completed.

The surrogate is defined as a $C^{1}$-continuous, positively homogeneous polynomial-type yield function composed of terms of degrees 2, 4, 6, and 8 \cite{Soare2008PolynomialYield,SoareBarlat2010ConvexPolynomial}:
\begin{equation}
    \bar{\sigma}_{\mathrm{poly}}(\boldsymbol{\sigma})
    =
    \sum_{m\in\{2,4,6,8\}}
    \left[
    \sum_{i+j+k=m}
    A_{ijk}^{(m)}
    \sigma_{xx}^{i}
    \sigma_{yy}^{j}
    \sigma_{xy}^{k}
    \right]^{1/m},
    \label{eq:polynomial_surrogate}
\end{equation}
where $A_{ijk}^{(m)}$ are polynomial coefficients. The fitting samples are generated by sampling stress directions $\{\mathbf{s}_{\iota}\}_{\iota=1}^{N_{s}}$ on the unit sphere in the three-component plane-stress space $(\sigma_{xx},\sigma_{yy},\sigma_{xy})$ and scaling each direction to the discovered neural yield level set. Using the positive homogeneity of the neural yield function, the corresponding stress sample is
\begin{equation}
    \boldsymbol{\sigma}_{\iota}
    =
    \frac{\sigma_{\mathrm{Y}}}
    {\bar{\sigma}_{\boldsymbol{\theta}}(\mathbf{s}_{\iota})}
    \mathbf{s}_{\iota},
    \qquad
    \bar{\sigma}_{\boldsymbol{\theta}}(\boldsymbol{\sigma}_{\iota})
    =
    \sigma_{\mathrm{Y}} .
    \label{eq:polynomial_sampling}
\end{equation}
Here, $\sigma_{\mathrm{Y}}=R(0)$ denotes the initial yield stress level used for the surrogate fit.
In this fitting step, $N_{s}=100{,}000$ stress samples are used, and the polynomial coefficients are obtained from the mean-squared-error (MSE) loss
\begin{equation}
    \mathcal{L}_{\mathrm{MSE}}
    =
    \frac{1}{N_{s}}
    \sum_{\iota=1}^{N_{s}}
    \left[
    \bar{\sigma}_{\mathrm{poly}}(\boldsymbol{\sigma}_{\iota})
    -
    \bar{\sigma}_{\boldsymbol{\theta}}(\boldsymbol{\sigma}_{\iota})
    \right]^{2}.
    \label{eq:polynomial_fitting_loss}
\end{equation}
This fitting step is performed after the physics-informed discovery problem has been solved. The neural yield function is first identified from displacement and reaction force data, and the polynomial surrogate is then fitted to that identified function.

The polynomial surrogate does not replace the mechanically constrained neural yield function used during discovery. Unlike the convex neural network, the polynomial surrogate is not convex by construction and introduces an additional approximation error that requires assessment before deployment. The two representations therefore serve distinct purposes in the framework. The convex neural network representation serves as the constrained search space during inverse identification, whereas the polynomial surrogate provides a smooth analytical approximation for conventional implicit finite element implementation.

\section{Results and Validation}
\label{sec:results}

\begin{figure}[!htbp]
\centering
\includegraphics[width=\textwidth]{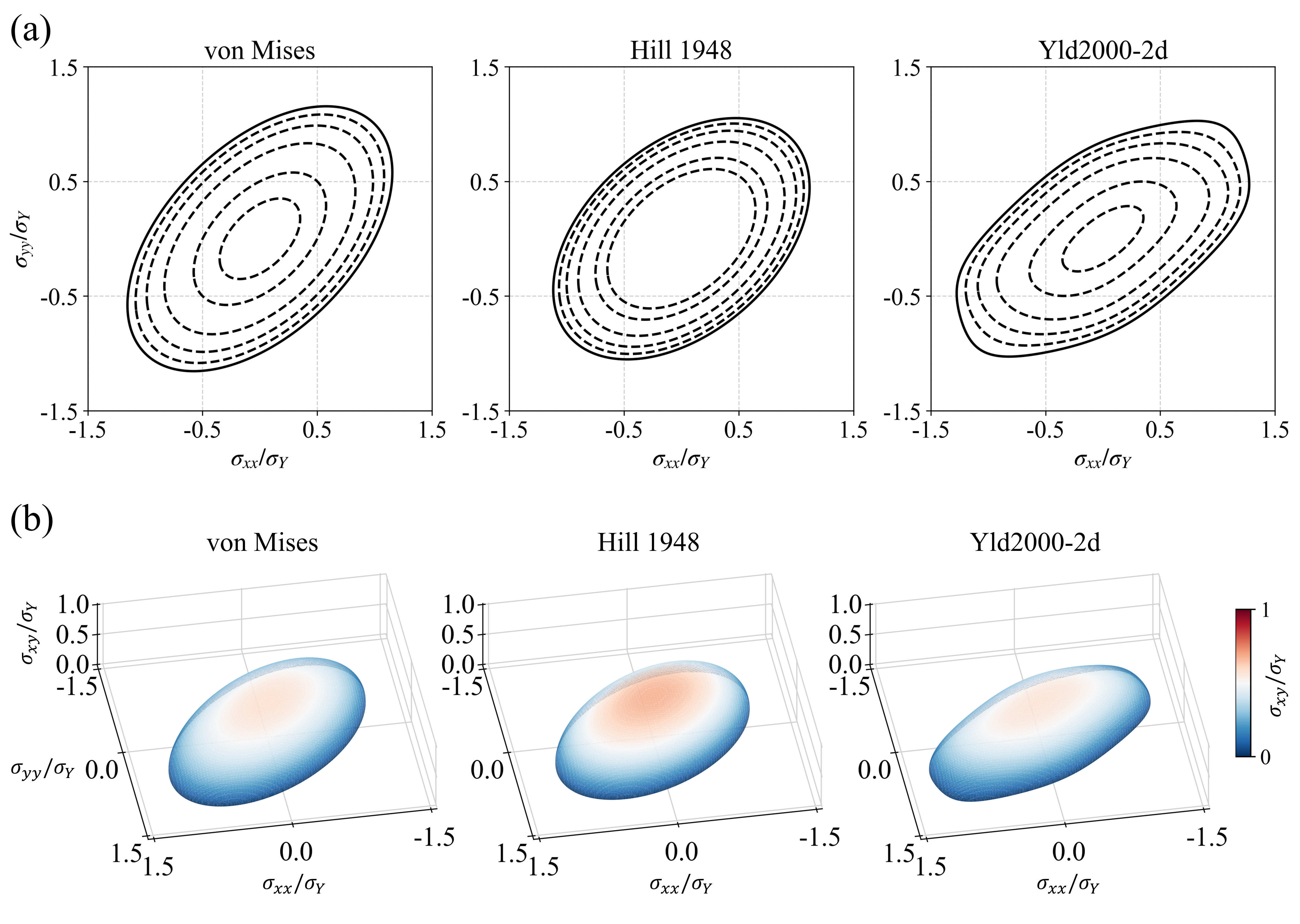}
\caption{
Three benchmark yield functions.
(a) Two-dimensional yield contours of the von Mises, Hill 1948, and Yld2000-2d models on the $\sigma_{xx}$--$\sigma_{yy}$ plane for prescribed shear stress levels normalized by the initial yield stress, $\sigma_{xy}/\sigma_{\mathrm{Y}} = 0.0$, $0.2$, $0.3$, $0.4$, $0.5$, and $0.55$.
(b) Corresponding three-dimensional yield surfaces in normalized stress space, where the color denotes $\sigma_{xy}/\sigma_{\mathrm{Y}}$.
}
\label{fig:benchmark_yield_functions}
\end{figure}

The proposed framework is validated using three benchmark yield functions, namely von Mises, Hill 1948, and Yld2000-2d, as shown in \figref{fig:benchmark_yield_functions}. These benchmarks represent isotropic, quadratic anisotropic, and non-quadratic anisotropic yielding, respectively, with known ground truth for direct comparison. Their definitions and parameters are given in Appendix A. In \figref{fig:benchmark_yield_functions}(a), the two-dimensional contours are shown at multiple shear stress levels normalized by the initial yield stress $\sigma_{\mathrm{Y}}=R(0)$, while \figref{fig:benchmark_yield_functions}(b) shows the corresponding three-dimensional yield surfaces in the plane-stress space.

For each benchmark, validation data are generated by FE simulations under plane-stress conditions using Abaqus with a user-defined material subroutine and implicit backward Euler return mapping. The two-void specimen in \figref{fig:problem_setting} is discretized by 4,485 nodes and 8,576 linear triangular plane-stress elements, and the same mesh is used in the inverse analysis. The four displacement-controlled loading cases prescribe boundary displacements corresponding to UTx, UTy, BT, and SS, with displacement amplitude $\delta=0.5~\mathrm{mm}$.

The elastic constants are set to $E=69~\mathrm{GPa}$ and $\nu=0.3$, and the known isotropic hardening response is prescribed in MPa as
\begin{equation}
    R(\bar{\varepsilon}^{p})
    =
    417.501\left(0.00457+\bar{\varepsilon}^{p}\right)^{0.22194}.
    \label{eq:validation_hardening_law}
\end{equation}
Each loading case is simulated for 500 load increments, and displacement fields and reaction force data are extracted every 5 load increments, resulting in 100 training snapshots per loading case. This temporal subsampling reflects the practical situation in which full-field measurements may be available only at finite acquisition intervals rather than at every load increment of a numerical simulation. The inverse analysis uses only the sampled displacement fields and reaction force data as inverse inputs; stress fields, plastic strain fields, and ground truth yield surface data are reserved for validation and subsequent interpretation. The explicit forward Euler differentiable stress update in Section~\ref{subsec:physics_informed_training} is then applied with $\mathcal{G}_{c}=\{4,8,12,\ldots,100\}$, and the sharpness schedule uses the transition-width parameter $\zeta=1/\beta$ introduced in Algorithm~\ref{alg:sharpness_schedule}. The main validation settings are summarized in \tabref{tab:validation_training_settings}.

\begin{table}[!htbp]
\centering
\caption{Settings used for the benchmark validation studies.}
\label{tab:validation_training_settings}
\begin{tabular}{lll}
\toprule
Quantity & Symbol & Setting \\
\midrule
Mesh resolution &  & 4,485 nodes; 8,576 elements \\
Displacement amplitude & $\delta$ & $0.5~\mathrm{mm}$ \\
Loading cases & $\mathcal{C}$ & UTx, UTy, BT, SS \\
Number of load increments &  & 500 per loading case \\
Sampling interval &  & every 5 increments \\
Number of training snapshots &  & 100 per loading case \\
Training epochs & $N_{\mathrm{epoch}}$ & 2,000 \\
Neural network architecture &  & 4 layers; 32 units per layer \\
Optimization algorithm &  & Adam \\
Loss weights & $w_{c}$ & $w_{c}=1.0$ for all $c\in\mathcal{C}$ \\
Initial numerical sharpness & $\beta_{0}$ & 4 \\
Sharpness update factor & $\eta_{\zeta}$ & 0.9 for $\zeta=1/\beta$ \\
Sharpness update period & $N_{\zeta}$ & 200 epochs \\
Selected higher-order differentiation increments & $\mathcal{G}_{c}$ & $\{4,8,12,\ldots,100\}$ \\
\bottomrule
\end{tabular}
\end{table}

\FloatBarrier

\subsection{Discovery Results under Noise-Free Conditions}
\label{subsec:noise_free_discovery}

\begin{figure}[!htbp]
    \centering
    \includegraphics[width=\textwidth]{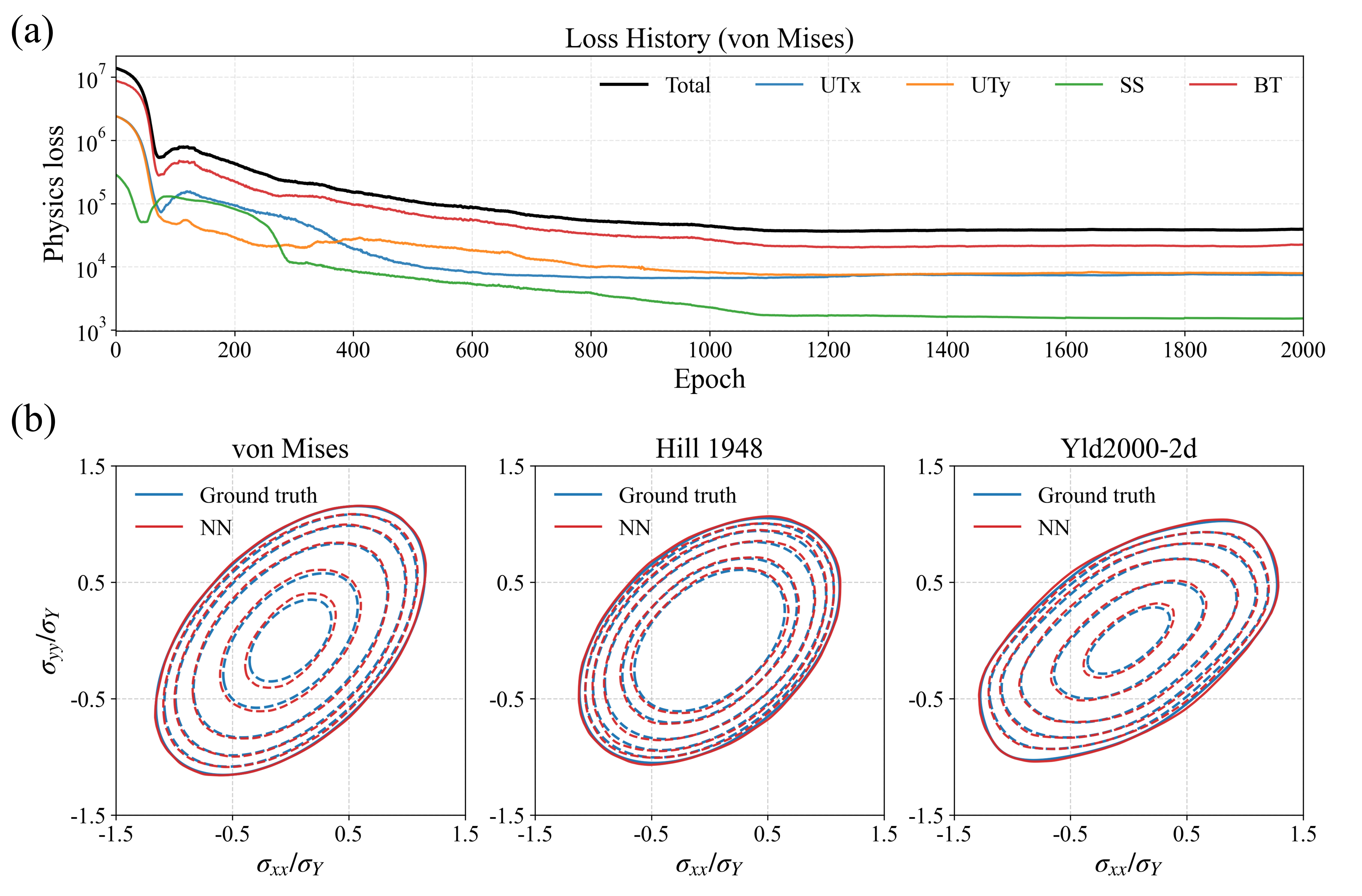}
    \caption{
    Yield function discovery results under noise-free conditions.
    (a) Loss history for the von Mises benchmark, showing the force equilibrium loss and the individual contributions from the four loading cases.
    (b) Comparison between the ground truth yield contours and contours predicted by the neural yield function for the three benchmark yield functions.
    }
    \label{fig:noise_free_discovery}
\end{figure}

The analysis first considers noise-free data to isolate the inverse identification problem from measurement perturbations. \figref{fig:noise_free_discovery}(a) shows a representative training history for the von Mises benchmark, where the total force equilibrium loss decreases and then approaches a plateau. The loss contributions from UTx, UTy, BT, and SS have different magnitudes because the four boundary conditions generate different internal force levels and stress states. They are nevertheless minimized through the same neural yield function, so the optimization is constrained by all loading cases simultaneously rather than by a single deformation mode.

\figref{fig:noise_free_discovery}(b) compares the inferred yield contours with the ground truth contours after training. For the von Mises benchmark, the neural yield function approximates the nearly circular contour family expected for isotropic yielding. For Hill 1948 and Yld2000-2d, the inferred contours also follow the anisotropic contour shapes over the shear stress levels considered, including the changes in curvature and orientation that distinguish the two anisotropic benchmarks. The comparison is obtained without stress observations, plastic strain measurements, direct yield surface data, or the closed-form benchmark equations; the yield function is inferred from displacement fields and reaction force data through the force equilibrium loss.

\begin{figure}[!htbp]
    \centering
    \includegraphics[width=\textwidth]{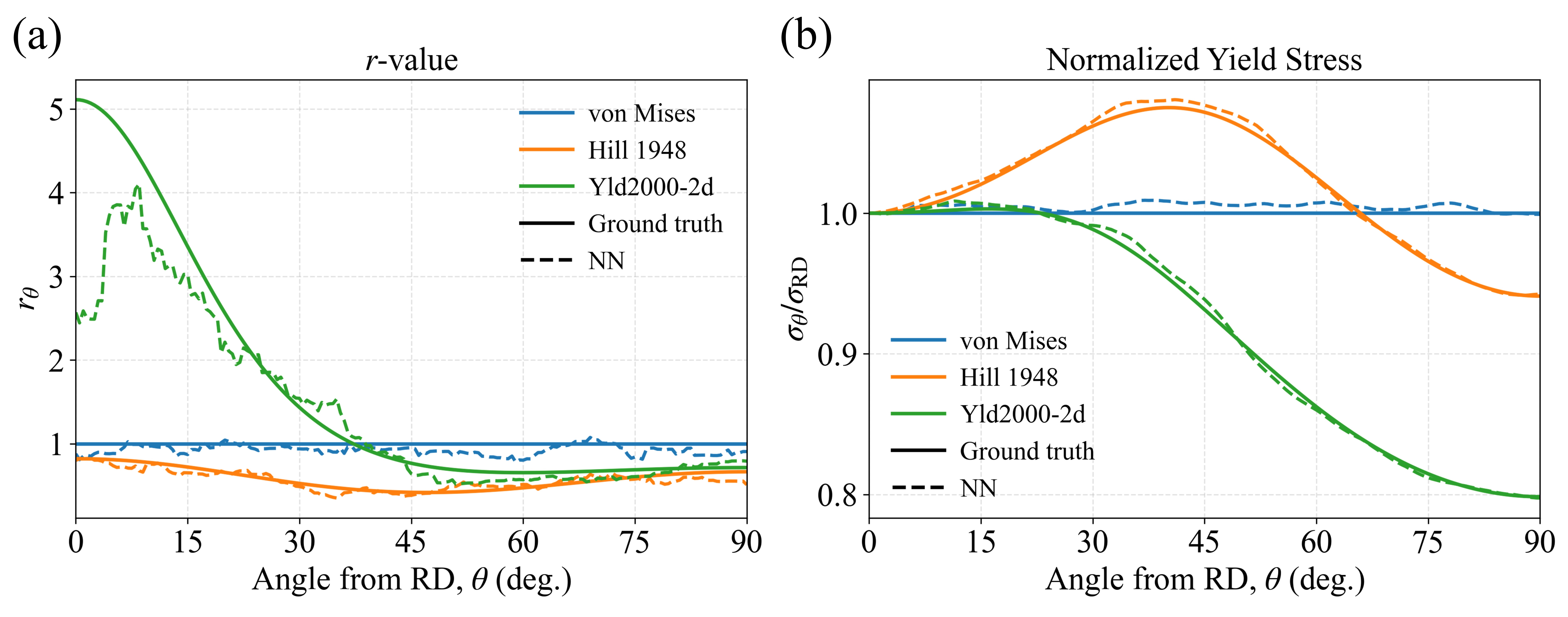}
    \caption{
    Directional anisotropy of the discovered yield functions under noise-free conditions.
    (a) $r$-values and (b) yield stresses normalized by the rolling-direction yield stress, $\sigma_{\theta}/\sigma_{\mathrm{RD}}$, evaluated as functions of the loading angle measured from the rolling direction.
    Solid lines denote the ground truth yield functions, whereas dashed lines denote predictions from the neural yield function.
    }
    \label{fig:directional_anisotropy}
\end{figure}

\figref{fig:directional_anisotropy} presents the same discovery results through directional quantities that are commonly used to characterize sheet-metal anisotropy. The $r$-values in \figref{fig:directional_anisotropy}(a) reproduce the overall ground truth trends but contain more local nonsmoothness, especially for the Yld2000-2d benchmark. This behavior is expected because the $r$-value depends on the derivative of the yield function with respect to stress through the AFR, whereas the convex neural network representation is treated as a $C^{0}$, piecewise differentiable function. The normalized yield stresses in \figref{fig:directional_anisotropy}(b) follow the ground truth trends more smoothly for the three benchmarks, including the directional variation of the anisotropic cases. These quantities describe the inferred yield surface as a function of loading angle in the uniaxial directions commonly used for sheet-metal anisotropy. The comparison therefore supports the contour-level results in \figref{fig:noise_free_discovery}(b), while also showing that derivative-based anisotropy measures such as $r$-values are more sensitive to the piecewise differentiable nature of the neural yield function than yield-stress trends.

\FloatBarrier

\subsection{Sensitivity to Measurement Noise}
\label{subsec:noise_sensitivity}

The displacement-noise study introduces perturbations in the measured displacement fields. Because the stress update is driven by strain increments obtained from displacement fields, small spatial perturbations in the measured displacement data can be amplified through differentiation before entering the force equilibrium loss. As detailed in Appendix B, zero-mean spatiotemporally correlated Gaussian noise is added as
\begin{equation}
u_{\text{noisy},d}(\boldsymbol{x},t)
=
u_{\text{clean},d}(\boldsymbol{x},t)
+
\sigma_{\text{noise}}\eta_{\text{norm},d}(\boldsymbol{x},t),
\qquad d\in\{x,y\},
\label{eq:noise_model_results}
\end{equation}
where $\eta_{\text{norm},d}$ is the normalized correlated noise field. The noise magnitudes are $\sigma_{\text{noise}}=0.1\,\mu$m, $0.2\,\mu$m, and $0.3\,\mu$m. These amplitudes are controlled sub-micrometer perturbations intended to emulate modern DIC measurement uncertainty in the displacement fields \cite{Sutton2009DIC,Hansen2021SRDIC,Siebert2021DICUncertainty}.

\begin{figure}[!htbp]
    \centering
    \includegraphics[width=\textwidth]{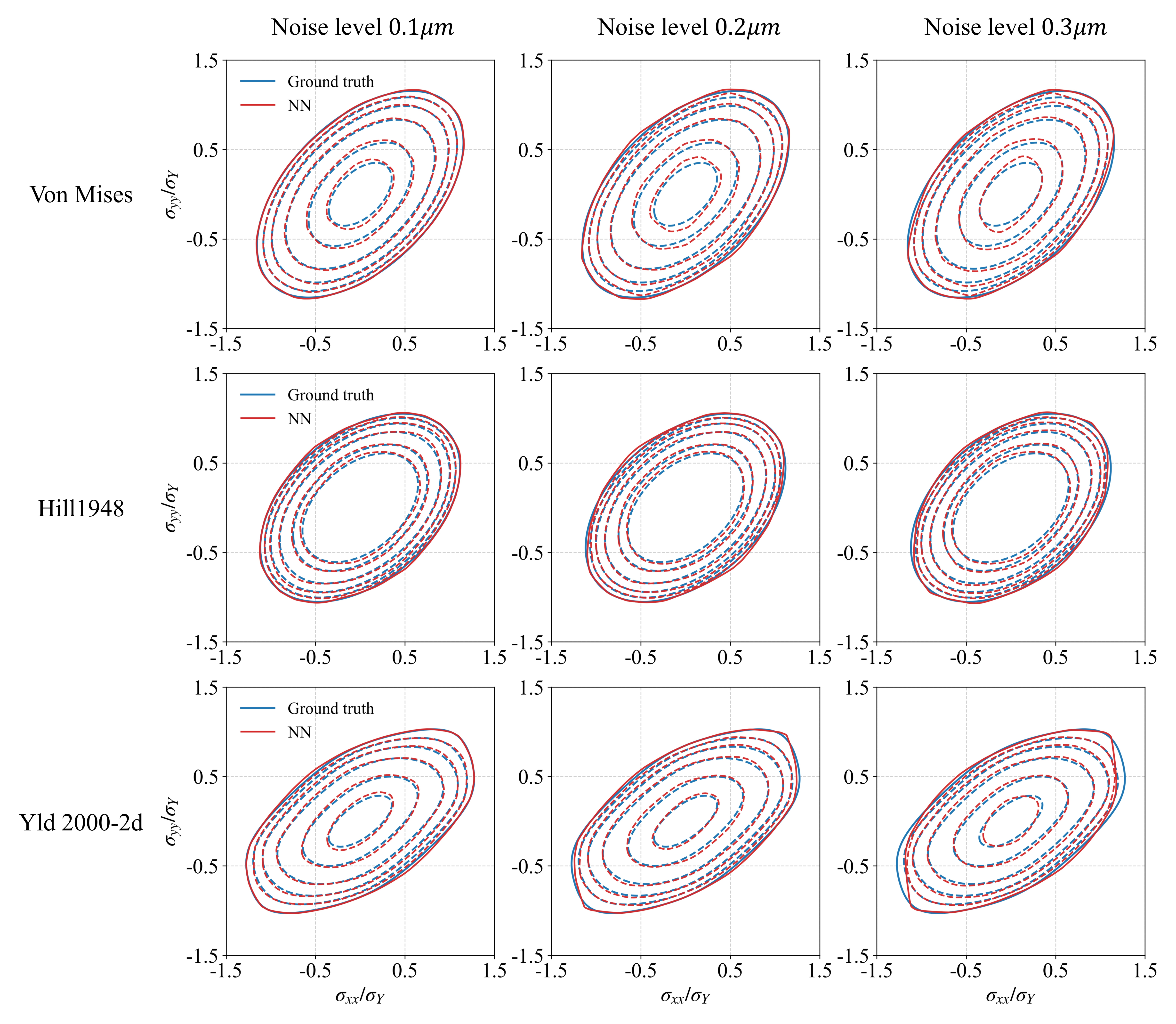}
    \caption{
    Sensitivity of yield function discovery to measurement noise.
    Ground truth yield contours and contours predicted by the neural yield function are compared for the three benchmark yield functions under displacement noise levels of $0.1\,\mu$m, $0.2\,\mu$m, and $0.3\,\mu$m.
    Rows correspond to the von Mises, Hill 1948, and Yld2000-2d models, while columns correspond to increasing noise levels.
    }
    \label{fig:noise_sensitivity}
\end{figure}

\figref{fig:noise_sensitivity} compares the ground truth and predicted yield contours for the three benchmarks, with rows corresponding to the benchmark yield functions and columns corresponding to increasing noise levels. The predicted contours remain consistent with the ground truth contours over the tested noise range, although the contours become progressively more distorted and deviate further from the ground truth as the displacement noise level increases. The von Mises benchmark serves as a relatively simple isotropic reference case, whereas the Hill 1948 and Yld2000-2d benchmarks require the directional dependence to be retained under noisy displacement-derived strain increments.

In \figref{fig:noise_sensitivity}, the degradation is gradual rather than a sudden loss of contour agreement. At $0.1\,\mu$m, the predicted contours remain close to the ground truth contours for all three benchmarks. At $0.2\,\mu$m and $0.3\,\mu$m, the differences are more visible near the outer contours and in the anisotropic cases, where the curvature and directional stretching of the yield surface place stronger demands on the displacement-derived strain field. The trend is consistent with the structure of the inverse problem because the data do not provide direct yield surface supervision, and the force equilibrium residual constrains the yield function through the stress fields generated by noisy displacement inputs. The noise study therefore motivates the identifiability analysis that follows, where the role of the stress states generated by the loading cases is analyzed more directly.

\FloatBarrier

\subsection{Identifiability Analysis}
\label{subsec:identifiability_analysis}

\begin{figure}[!htbp]
    \centering
    \includegraphics[width=\textwidth]{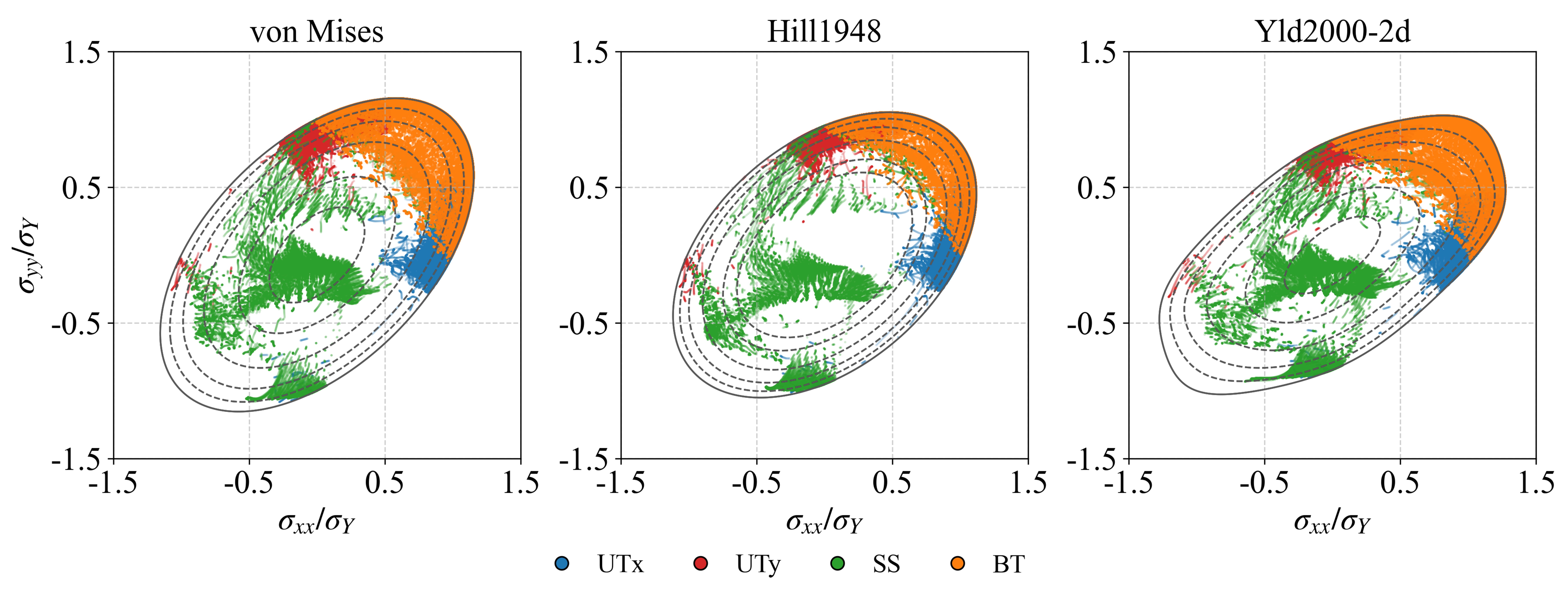}
    \caption{
    Plastically active stress states obtained from ground truth FE simulations.
    For each benchmark yield function, the points represent stress states at integration points where plastic yielding occurs, accumulated over the sampled loading history and projected onto the corresponding ground truth yield surface.
    Colors denote the four loading cases, UTx, UTy, BT, and SS.
    These distributions show which regions of the yield surface are sampled by the stress states generated under the prescribed loading cases and therefore which regions are indirectly constrained by the available data.
    }
    \label{fig:stress_state_coverage}
\end{figure}

The yield contour comparisons do not by themselves indicate which regions of stress space are directly constrained by the force equilibrium residual. \figref{fig:stress_state_coverage} therefore shows the stress states at plastically active integration points from the ground truth FE simulations. The points are grouped by loading case, accumulated over the sampled loading history, and projected onto the corresponding ground truth yield surface. These stress states are used only for interpretation and are not used by the discovery algorithm during training.

The four loading cases in \figref{fig:stress_state_coverage} generate distinct distributions of plastically active stress states. UTx, UTy, BT, and SS contribute different combinations of normal and shear stresses, and their combined distribution covers multiple regions of each benchmark yield surface. Regions that are not visited by plastically active stress states provide weaker force equilibrium constraints on the corresponding part of the yield surface. In those regions, the inferred shape is influenced more strongly by the convex neural network representation, the assumed symmetry, and the adopted homogeneity condition.

\begin{figure}[!htbp]
    \centering
    \includegraphics[width=\textwidth]{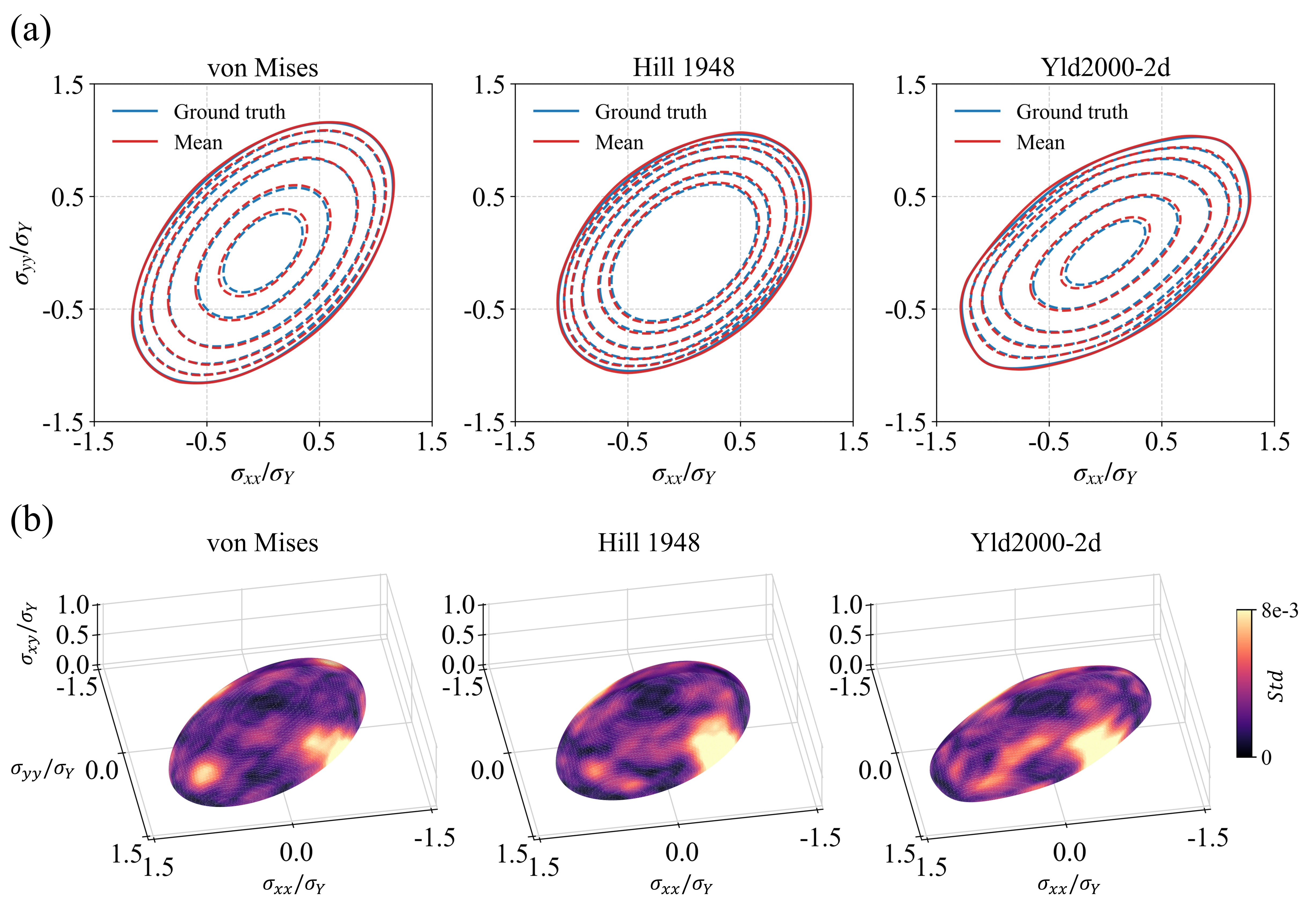}
    \caption{
    Deep-ensemble epistemic uncertainty for the discovered yield functions.
    For each benchmark yield function, yield function discovery was repeated 10 times with different random initializations.
    (a) Mean yield contours obtained from the 10 discovery results, compared with the corresponding ground truth yield contours.
    (b) Standard deviation of the neural yield function predictions evaluated on the corresponding ground truth yield surface in the 3D stress space.
    }
    \label{fig:epistemic_uncertainty}
\end{figure}

\figref{fig:epistemic_uncertainty} adds an uncertainty-based interpretation through a deep ensemble constructed from 10 independently initialized optimizations. The mean contours in \figref{fig:epistemic_uncertainty}(a) remain consistent with the ground truth contours, indicating that the result is not specific to one initialization. The standard-deviation fields in \figref{fig:epistemic_uncertainty}(b) then show where the inferred yield surface varies across the ensemble. Regions sampled densely by plastically active stress states in \figref{fig:stress_state_coverage} generally show lower epistemic uncertainty. By contrast, the opposite-sign normal-stress regions, characterized by $\sigma_{xx}\sigma_{yy}<0$, are sparsely represented by the prescribed loading cases; the absence of stress states in these regions is consistent with the elevated epistemic uncertainty in the corresponding regions of \figref{fig:epistemic_uncertainty}. Together, \figref{fig:stress_state_coverage} and \figref{fig:epistemic_uncertainty} indicate that the constraint on the inferred yield surface depends not only on the neural yield function but also on whether the loading cases generate plastically active stress states over the relevant regions of the yield surface.

The uncertainty analysis is restricted to epistemic uncertainty estimated from noise-free data through the deep ensemble. It does not quantify aleatoric uncertainty associated with noisy displacement measurements or uncertain experimental conditions. Treating both epistemic and aleatoric uncertainty in noisy inverse identification from full-field displacement data would require a more complete probabilistic or Bayesian framework, which is beyond the scope of the present study and remains a topic for future work.

\FloatBarrier

\subsection{Finite Element Deployment}
\label{subsec:fem_deployment}

Deployment of the discovered yield function in an implicit finite element setting requires a smooth analytical form, because conventional implicit finite element stress integration is not directly suited to the piecewise differentiable convex neural network representation used during discovery. The polynomial surrogate is therefore evaluated as a smooth approximation of the discovered neural yield function for an implicit finite element simulation. The Yld2000-2d benchmark is used as a representative case because it has the most complex non-quadratic anisotropic yield surface among the three benchmarks.

\begin{figure}[!htbp]
    \centering
    \includegraphics[width=\textwidth]{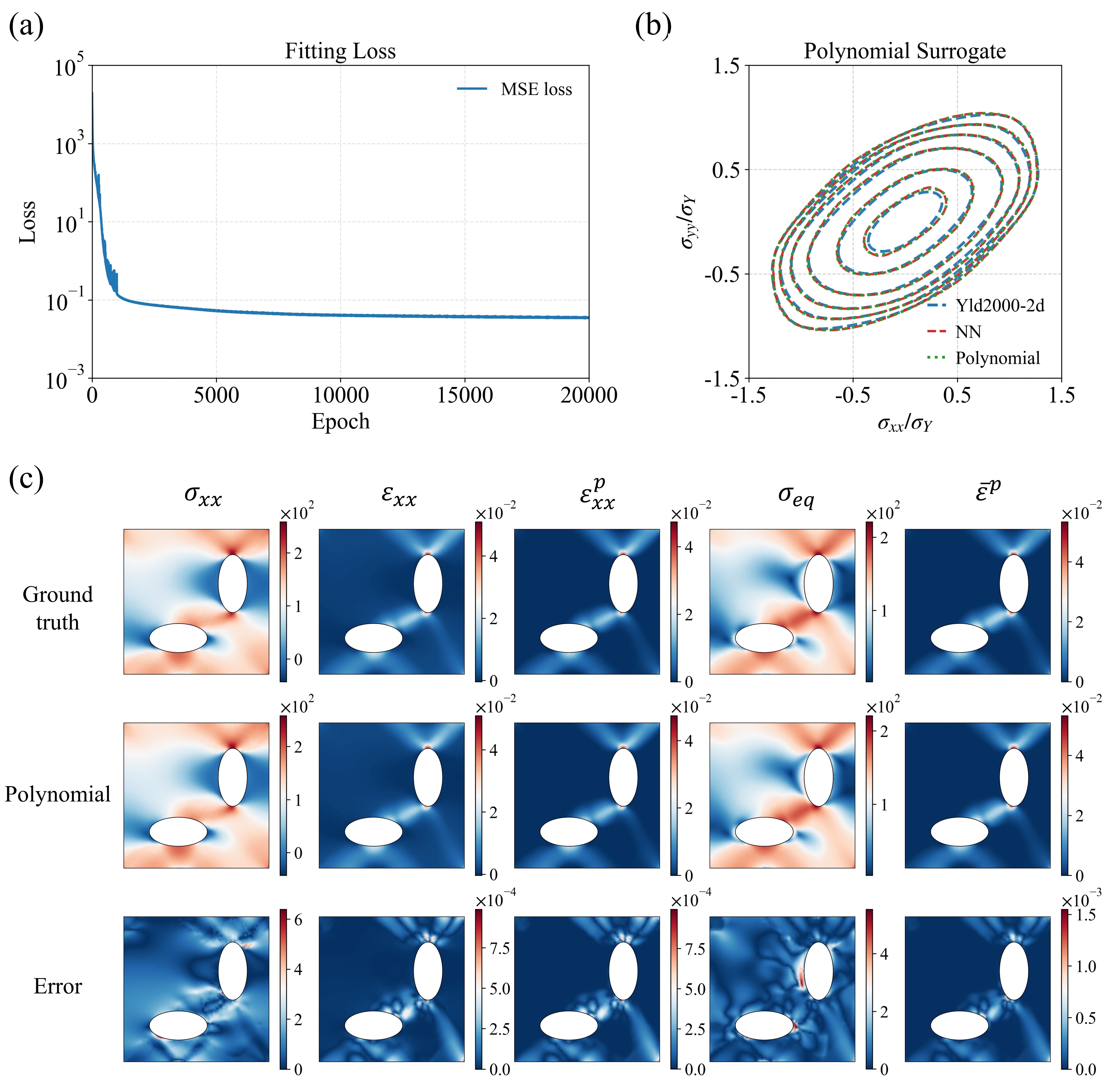}
    \caption{
    Representative polynomial surrogate modeling for finite element deployment.
    (a) Fitting loss history for approximating the discovered neural yield function with a polynomial yield function.
    (b) Yield contour comparison among the ground truth Yld2000-2d model, the discovered neural yield function, and the fitted polynomial surrogate.
    (c) Finite element simulation results obtained using the ground truth yield function and the fitted polynomial surrogate, along with the corresponding absolute error fields.
    The field quantities include $\sigma_{xx}$, $\varepsilon_{xx}$, $\varepsilon_{xx}^{p}$, $\sigma_{\mathrm{eq}}$, and $\bar{\varepsilon}^{p}$.
    }
    \label{fig:polynomial_surrogate_deployment}
\end{figure}

As described in Section~\ref{subsec:polynomial_surrogate}, a polynomial surrogate is fitted to 100,000 stress samples on the level set of the discovered neural yield function. \figref{fig:polynomial_surrogate_deployment}(a) presents the mean-squared-error fitting loss, which decreases rapidly at early epochs and then approaches a plateau. \figref{fig:polynomial_surrogate_deployment}(b) compares the ground truth Yld2000-2d contours, the discovered neural yield function, and the fitted polynomial surrogate. The polynomial contours remain consistent with the neural contours and with the ground truth contours over the shear stress levels considered, indicating that the surrogate fit retains the contours relevant to this benchmark comparison.

\figref{fig:polynomial_surrogate_deployment}(c) compares finite element fields obtained with the ground truth yield function and with the fitted polynomial surrogate for the UTx loading case at the final loading increment. The field quantities include $\sigma_{xx}$, $\varepsilon_{xx}$, $\varepsilon_{xx}^{p}$, $\sigma_{\mathrm{eq}}$, and $\bar{\varepsilon}^{p}$, so the comparison covers both stress and accumulated plastic response. The surrogate-based simulation produces similar spatial distributions for the stress and plastic-strain fields shown in \figref{fig:polynomial_surrogate_deployment}(c), while the absolute error fields identify localized discrepancies near the heterogeneous deformation zones. The comparison illustrates a two-stage use of the framework, in which a mechanically constrained neural yield function is first discovered from displacement and reaction force data and then approximated by a smooth polynomial surrogate for finite element deployment. The result is therefore a representative deployment assessment for the tested loading case and selected polynomial approximation.

\FloatBarrier

\section{Discussion}
\label{sec:discussion}

The benchmark studies support the central claim of the proposed framework: anisotropic yield functions can be represented by a convex neural yield function and identified from full-field displacement data and reaction force data through force equilibrium. In the tested cases, the force equilibrium framework approximated the benchmark yield contours and followed directional anisotropy trends without stress observations, plastic strain measurements, direct yield surface data, or the closed-form benchmark equations. The significance of this result is that the neural yield function is inferred as the unknown constitutive component inside elastoplastic stress integration, rather than as a supervised contour fit in stress space. The neural yield function is converted into internal force predictions through the stress update and constrained by force equilibrium across multiple loading cases, so the identification problem is formulated in terms of observable mechanical data.

The convex neural network representation constrains the constitutive search space for this inverse problem. The network represents a scalar yield function rather than an unconstrained stress-strain map, and the architecture enforces convexity and positive homogeneity of degree one. The assumed tension-compression symmetry is imposed by symmetrization. Embedding this representation in a differentiable elastoplastic stress update allows the force equilibrium residual to train the yield function while preserving the mechanical structure of rate-independent associated plasticity. This distinction separates the framework from supervised yield surface learning or stress-update surrogates \cite{Nascimento2023YieldSurfaces,FazilyYoon2023StressIntegration,Fuhg2023ModularElastoplasticity,Fuhg2023EnhancingYield}. The framework is nevertheless not a complete constitutive discovery method: the neural yield function is treated as $C^{0}$ and piecewise differentiable, the hardening law is prescribed, and the AFR is assumed. With these quantities prescribed, the study isolates whether the yield function can be inferred from displacement and force data before broader coupled identification problems are attempted.

The validation results clarify the conditions under which the discovered yield function is well constrained. The noise-free results show the yield surface approximation attainable under idealized displacement data, and the displacement-noise studies show the gradual degradation induced by modern DIC-scale perturbations. The identifiability analysis provides a complementary interpretation by clarifying how plastically active stress states enter and constrain the inverse problem. Stress states generated by the specimen and loading cases enter the force equilibrium residual through the stress update, whereas regions sparsely represented by those stress states are governed more strongly by the constrained function class. The elevated epistemic uncertainty in the opposite-sign normal-stress regions, $\sigma_{xx}\sigma_{yy}<0$, is consistent with the lack of plastically active stress states in those regions. This behavior indicates where the inferred yield surface is supported by the generated stress states and where additional loading cases or a redesigned specimen would be needed.

The polynomial surrogate study extends the framework toward finite element deployment. The convex neural network is well suited for the constrained inverse-identification stage, but its $C^{0}$, piecewise differentiable form is not directly suited to conventional implicit return mapping. Fitting a smooth polynomial-type yield function to the discovered neural yield function enables standard finite element implementation. The Yld2000-2d deployment comparison shows that this post-processing step can retain the relevant yield contours in the tested case. This comparison supports the surrogate for the tested benchmark case and selected polynomial approximation, while broader surrogate validity requires additional loading cases and surrogate forms.

Several limitations remain. The validation uses data generated by FE simulations rather than experimental DIC measurements; the elastic response and hardening law are prescribed, the AFR is assumed, and plane-stress conditions are imposed. With these assumptions, the present study does not address hardening characterization, cases governed by a non-associated flow rule, experimental boundary-condition uncertainty, or out-of-plane stress effects. Future work can extend the framework toward experimental DIC studies, richer specimen and loading design, simultaneous identification of yield and hardening behavior, three-dimensional stress states, and yield function representations compatible with implicit finite element solvers. In experimental extensions, practically realizable specimen geometries and displacement-controlled loading configurations should be designed to obtain full-field displacement measurements from DIC and global reaction-force data from mechanical testing, while accounting for practical uncertainty sources such as measurement noise, fixture compliance, alignment error, boundary-condition mismatch, and local material heterogeneity. The results further indicate that specimen design and loading selection directly affect the stress states entering the force equilibrium constraints and, consequently, the regions of the yield surface that can be inferred with confidence.

\section{Conclusions}
\label{sec:conclusions}

This study presented a physics-informed framework for discovering anisotropic yield functions from full-field displacement data and reaction force data. The main result is that the yield function can be identified as a mechanically constrained constitutive component inside elastoplastic stress integration, rather than through direct stress-space supervision. The yield function was represented by a convex neural network that enforces convexity and positive homogeneity of degree one while embedding the assumed tension-compression symmetry. A differentiable elastoplastic training procedure then coupled this neural yield function to force equilibrium, without using stress observations, plastic strain measurements, direct yield surface data, or a prescribed analytical yield function.

The framework was evaluated using the von Mises, Hill 1948, and Yld2000-2d benchmark yield functions. Under noise-free conditions, the discovered yield functions agreed qualitatively with the ground truth and followed the directional anisotropy trends. Displacement-noise studies showed gradual degradation as the noise level increased, while the inferred contours remained consistent with the ground truth yield contours over the tested levels. The identifiability and epistemic uncertainty analyses further showed that regions sampled by plastically active stress states were more strongly constrained, whereas sparsely represented opposite-sign normal-stress regions exhibited higher epistemic uncertainty. A smooth polynomial surrogate fitted to the discovered neural yield function was also evaluated in a representative Yld2000-2d finite element deployment case.

The current study is limited to benchmark data generated by FE simulations with prescribed elasticity and hardening, an assumed AFR, and plane-stress conditions. Future work can extend the framework toward experimental DIC data, richer specimen and loading design, simultaneous identification of yield and hardening behavior, three-dimensional stress states, and yield function representations compatible with implicit finite element solvers.

\clearpage

\section*{Declaration of Competing Interest}
The authors declare that they have no known competing financial interests or personal relationships that could have appeared to influence the work presented in this paper.

\section*{Data Availability}
The data used in this study is available from the corresponding author upon reasonable request.

\section*{Code Availability}
The code used in this study is available from the corresponding author upon reasonable request.

\section*{Acknowledgment}
This research was supported by the Ministry of Food and Drug Safety of Korea (Grant No. RS-2023-00215667) and by the InnoCORE Program of the Ministry of Science and ICT (MSIT), Republic of Korea (Grant No. N10260002).

\clearpage

\appendix

\renewcommand{\thesection}{Appendix \Alph{section}}

\counterwithin{figure}{section}
\counterwithin{table}{section}
\counterwithin{algorithm}{section}
\counterwithin{equation}{section}

\renewcommand{\thefigure}{\Alph{section}\arabic{figure}}
\renewcommand{\thetable}{\Alph{section}\arabic{table}}
\renewcommand{\thealgorithm}{\Alph{section}\arabic{algorithm}}
\renewcommand{\theequation}{\Alph{section}.\arabic{equation}}

\makeatletter
\def\fnum@table{Table~\thetable}
\def\fnum@figure{Figure~\thefigure}
\makeatother

\section{Benchmark Yield Functions}

The benchmark yield functions considered in this study are summarized.

\textbf{von Mises isotropic model.}
The von Mises equivalent stress is defined as
\begin{equation}
\bar{\sigma}(\boldsymbol{\sigma}) =
\sqrt{\sigma_{xx}^2 + \sigma_{yy}^2 
- \sigma_{xx}\sigma_{yy} + 3\sigma_{xy}^2
}.
\end{equation}

\vspace{0.75em}

\textbf{Hill 1948 anisotropic quadratic model.}
The Hill 1948 equivalent stress is expressed as
\begin{equation}
\bar{\sigma}(\boldsymbol{\sigma}) =
\sqrt{
\frac{1}{2}
\left[
(G_{\mathrm{Hill}}+H_{\mathrm{Hill}})\sigma_{xx}^2
+ (F_{\mathrm{Hill}}+H_{\mathrm{Hill}})\sigma_{yy}^2
- 2H_{\mathrm{Hill}}\sigma_{xx}\sigma_{yy}
+ 2N_{\mathrm{Hill}}\sigma_{xy}^2
\right]
}.
\end{equation}

The anisotropic coefficients used in this study are listed in \tabref{tab:hill}.

\begin{table}[h]
\centering
\caption{Anisotropic coefficients of Hill 1948 (AA6022-T4E32)}
\label{tab:hill}
\begin{tabular*}{0.65\linewidth}{@{\extracolsep{\fill}}cccc}
\toprule
$F_{\mathrm{Hill}}$ & $G_{\mathrm{Hill}}$ & $H_{\mathrm{Hill}}$ & $N_{\mathrm{Hill}}$ \\
\midrule
1.3251 & 1.0730 & 0.8799 & 2.2014 \\
\bottomrule
\end{tabular*}
\end{table}

\vspace{0.75em}

\textbf{Yld2000-2d anisotropic non-quadratic model.}
The Yld2000-2d yield function is defined in terms of two linearly transformed plane-stress vectors, denoted here by $\boldsymbol{X}$ and $\boldsymbol{Y}$. 
The equivalent stress is given by
\begin{equation}
\bar{\sigma}(\boldsymbol{\sigma}) =
\left[
\frac{1}{2}
\left(
|X_1' - X_2'|^a
+ |2Y_1' + Y_2'|^a
+ |Y_1' + 2Y_2'|^a
\right)
\right]^{1/a},
\end{equation}
where $X_1'$ and $X_2'$ are the principal values associated with $\boldsymbol{X}$, and $Y_1'$ and $Y_2'$ are the principal values associated with $\boldsymbol{Y}$.
The transformed stresses are expressed as
\begin{equation}
\boldsymbol{X}
=
\begin{bmatrix}
L'_{11}\sigma_{xx}+L'_{12}\sigma_{yy} \\
L'_{21}\sigma_{xx}+L'_{22}\sigma_{yy} \\
L'_{33}\sigma_{xy}
\end{bmatrix},
\qquad
\boldsymbol{Y}
=
\begin{bmatrix}
L''_{11}\sigma_{xx}+L''_{12}\sigma_{yy} \\
L''_{21}\sigma_{xx}+L''_{22}\sigma_{yy} \\
L''_{33}\sigma_{xy}
\end{bmatrix}.
\end{equation}
The linear transformation coefficients are
\begin{equation}
\begin{aligned}
L'_{11} &= \frac{2\alpha_1}{3},
&
L'_{12} &= -\frac{\alpha_1}{3}, \\
L'_{21} &= -\frac{\alpha_2}{3},
&
L'_{22} &= \frac{2\alpha_2}{3}, \\
L'_{33} &= \alpha_7,
&
L''_{33} &= \alpha_8, \\
L''_{11} &= \frac{8\alpha_5-2\alpha_3-2\alpha_6+2\alpha_4}{9},
&
L''_{12} &= \frac{4\alpha_6-4\alpha_4-4\alpha_5+\alpha_3}{9}, \\
L''_{21} &= \frac{4\alpha_3-4\alpha_5-4\alpha_4+\alpha_6}{9},
&
L''_{22} &= \frac{8\alpha_4-2\alpha_6-2\alpha_3+2\alpha_5}{9}.
\end{aligned}
\end{equation}
For $\boldsymbol{X}=[X_{11},X_{22},X_{12}]^T$ and $\boldsymbol{Y}=[Y_{11},Y_{22},Y_{12}]^T$, the principal values are computed as
\begin{equation}
\begin{aligned}
X_1' &= \frac{X_{11}+X_{22}}{2}
+\sqrt{\left(\frac{X_{11}-X_{22}}{2}\right)^2+X_{12}^{2}},
&
X_2' &= \frac{X_{11}+X_{22}}{2}
-\sqrt{\left(\frac{X_{11}-X_{22}}{2}\right)^2+X_{12}^{2}}, \\
Y_1' &= \frac{Y_{11}+Y_{22}}{2}
+\sqrt{\left(\frac{Y_{11}-Y_{22}}{2}\right)^2+Y_{12}^{2}},
&
Y_2' &= \frac{Y_{11}+Y_{22}}{2}
-\sqrt{\left(\frac{Y_{11}-Y_{22}}{2}\right)^2+Y_{12}^{2}} .
\end{aligned}
\end{equation}
The anisotropic coefficients used for the Yld2000-2d model are summarized in \tabref{tab:yld2000}.

\begin{table}[h]
\centering
\caption{Anisotropic coefficients of Yld2000-2d (Ferritic stainless steel type 409)}
\label{tab:yld2000}
\begin{tabular*}{0.95\linewidth}{@{\extracolsep{\fill}}ccccccccc}
\toprule
$\alpha_1$ & $\alpha_2$ & $\alpha_3$ & $\alpha_4$ & $\alpha_5$ & $\alpha_6$ & $\alpha_7$ & $\alpha_8$ & $a$ \\
\midrule
0.9835 & 1.1182 & 0.7435 & 0.8517 & 0.8879 & 0.6511 & 0.9790 & 1.0810 & 6 \\
\bottomrule
\end{tabular*}
\end{table}

\section{Spatiotemporally Correlated Gaussian Noise}

To prescribe controlled noise in the displacement field, additive Gaussian noise with both spatial and temporal correlations is introduced.
The noise is generated in three steps: sampling uncorrelated Gaussian noise, applying temporal smoothing, and applying spatial smoothing.

Let $\bar{\eta}_{d}(\boldsymbol{x}, t) \sim \mathcal{N}(0,1)$ denote independent noise at spatial location $\boldsymbol{x}$, time step $t$, and displacement component $d\in\{x,y\}$.
Temporal correlation is introduced by applying a Gaussian kernel along the time axis,
\begin{equation}
\hat{\eta}_{d}(\boldsymbol{x}, t)
=
\sum_{\tau}
G_{\text{time}}(t - \tau;\, \sigma_{\text{time}})
\, \bar{\eta}_{d}(\boldsymbol{x}, \tau),
\end{equation}
where $\sigma_{\text{time}} = 5$ time steps. 
For each time step and displacement component, spatial correlation is introduced across nodes $\{\boldsymbol{x}_i\}_{i=1}^{m}$,
\begin{equation}
\eta_{d}(\boldsymbol{x}_i, t)
=
\sum_{j=1}^{m}
w_{ij}
\, \hat{\eta}_{d}(\boldsymbol{x}_j, t),
\end{equation}
where the weights are defined as
\begin{equation}
w_{ij}
=
\frac{
\exp\left(-\|\boldsymbol{x}_i - \boldsymbol{x}_j\|^2 / (2\sigma_{\text{space}}^2)\right)
}{
\sum_{k}
\exp\left(-\|\boldsymbol{x}_i - \boldsymbol{x}_k\|^2 / (2\sigma_{\text{space}}^2)\right)
},
\end{equation}
with $\sigma_{\text{space}} = 5~\mathrm{mm}$. 
The resulting noise is normalized at each time step and for each displacement component,
\begin{equation}
\eta_{\text{norm},d}(\boldsymbol{x}, t)
=
\frac{\eta_{d}(\boldsymbol{x}, t) - \mu_{td}}{\sigma_{td}},
\end{equation}
where $\mu_{td}$ and $\sigma_{td}$ are the mean and standard deviation of $\eta_{d}(\boldsymbol{x},t)$ over all spatial locations at time step $t$ and displacement component $d$.
The normalized noise field is then scaled by a prescribed magnitude,
\begin{equation}
\eta_{\text{final},d}(\boldsymbol{x}, t)
=
\sigma_{\text{noise}} \, \eta_{\text{norm},d}(\boldsymbol{x}, t).
\end{equation}
The noisy displacement field is then obtained as
\begin{equation}
u_{\text{noisy},d}(\boldsymbol{x}, t)
=
u_{\text{clean},d}(\boldsymbol{x}, t)
+
\eta_{\text{final},d}(\boldsymbol{x}, t).
\end{equation}

The smoothing parameters are set to $\sigma_{\text{time}} = 5$ time steps and $\sigma_{\text{space}} = 5~\mathrm{mm}$, which introduce finite spatial and temporal correlation without making the noise field spatially uniform. 
To evaluate noise sensitivity, three noise levels are considered,
\[
\sigma_{\text{noise}} \in \{0.1~\mu\mathrm{m},\, 0.2~\mu\mathrm{m},\, 0.3~\mu\mathrm{m}\},
\]
The three values prescribe controlled sub-micrometer displacement perturbations intended to emulate modern DIC measurement uncertainty, rather than calibrating the noise model to a specific experimental imaging system \cite{Sutton2009DIC,Hansen2021SRDIC,Siebert2021DICUncertainty}.
For clarity, the overall procedure is summarized in Algorithm~\ref{alg:noise}.

\begin{algorithm}[!htbp]
\caption{Generation of spatiotemporally correlated Gaussian noise}
\label{alg:noise}
\begin{algorithmic}[1]
\Statex \textbf{Input:} Nodal coordinates $\{\boldsymbol{x}_i\}_{i=1}^m$, clean displacement $\boldsymbol{u}_{\text{clean}}$, parameters $\sigma_{\text{time}}, \sigma_{\text{space}}, \sigma_{\text{noise}}$
\Statex \textbf{Output:} Noisy displacement field $\boldsymbol{u}_{\text{noisy}}$

\State Sample $\bar{\eta}_{d}(\boldsymbol{x}_i, t) \sim \mathcal{N}(0,1)$ for each component $d\in\{x,y\}$

\For{each node and component $d$}
\State $\hat{\eta}_{d}(\boldsymbol{x}_i, t) \gets \sum_{\tau} G_{\text{time}}(t-\tau)\,\bar{\eta}_{d}(\boldsymbol{x}_i, \tau)$
\EndFor

\For{each time step}
\For{each node and component $d$}
\State $\eta_{d}(\boldsymbol{x}_i, t) \gets \sum_{j} w_{ij}\,\hat{\eta}_{d}(\boldsymbol{x}_j, t)$
\EndFor
\EndFor

\For{each time step and component $d$}
\State $\eta_{\text{norm},d} \gets (\eta_{d} - \mu_{td})/\sigma_{td}$
\EndFor

\State $\eta_{\text{final},d} \gets \sigma_{\text{noise}}\,\eta_{\text{norm},d}$

\State $u_{\text{noisy},d} \gets u_{\text{clean},d} + \eta_{\text{final},d}$
\end{algorithmic}
\end{algorithm}

\clearpage
\begingroup
\setlength{\bibsep}{0pt}
\renewcommand{\baselinestretch}{0.9}\normalsize
\bibliography{main}
\endgroup

\end{document}